\documentclass[sigplan,nonacm,screen,pbalance,acmthm]{acmart}
\usepackage{multirow}
\usepackage{listings}
\usepackage{xcolor}
\usepackage{array}
\usepackage{amsthm}
\usepackage{cancel}

\usepackage[utf8]{inputenc} % allow utf-8 input
\usepackage{url}            % simple URL typesetting
\usepackage{amsfonts}       % blackboard math symbols
\usepackage{nicefrac}       % compact symbols for 1/2, etc.
\usepackage{xspace}
\usepackage[shortlabels]{enumitem}
\makeatletter
\def\url@leostyle{%
  \@ifundefined{selectfont}{\def\UrlFont{\sf}}{\def\UrlFont{\small\ttfamily}}}
\makeatother
\usepackage{cleveref}
\usepackage{subcaption}
\usepackage{mathpartir}

\usepackage[frozencache,cachedir=.]{minted}
\definecolor{codegreen}{rgb}{0,0.6,0}
\definecolor{codegray}{rgb}{0.5,0.5,0.5}
\definecolor{codepurple}{rgb}{0.58,0,0.82}
\definecolor{backcolour}{rgb}{0.95,0.95,0.92}

\lstdefinestyle{mystyle}{
    backgroundcolor=\color{white},   
    commentstyle=\color{blue},
    keywordstyle=[1]\color{blue},
    keywordstyle=[2]\color{magenta},
    keywordstyle=[3]\color{red},
    keywordstyle=[4]\color{codegray},
    numberstyle=\tiny\color{codegray},
    stringstyle=\color{codegreen},
    basicstyle=\ttfamily\footnotesize,
    breakatwhitespace=false,         
    breaklines=true,                 
    captionpos=b,                    
    keepspaces=true,                 
    numbers=left,                    
    numbersep=5pt,                  
    showspaces=false,                
    showstringspaces=false,
    showtabs=false,                  
    tabsize=2,
    texcl=true,
    escapeinside={(*@}{@*)}
}
\lstdefinelanguage{mlir}{
    alsoletter={\%,\#,!,.,_},
    morekeywords={\%},
    keywords=[1]{func, return, attributes},
    keywords=[2]{loop, \#tile, \#sum, slice, yield, atomic, spmd, range},
    keywords=[4]{tensor, f32},
    showstringspaces=false,
	breaklines=true,
    breakatwhitespace=true,
    morestring=[b]",
    morecomment=[l]{//},
}
\newcommand{\inlinecode}[1]{{\lstinline[basicstyle=\ttfamily]!#1!}}
\newcommand{\inlc}[1]{\inlinecode{#1}}

\newif\ifcomments
\commentstrue

\ifcomments
\newcommand{\authorcomment}[3]{\textcolor{#2}{#1: {\emph{#3}}}}
\else
\newcommand{\authorcomment}[3]{}
\fi

\restylefloat{figure}

\newcommand{\reducedstrut}{\vrule width 0pt height .9\ht\strutbox depth .9\dp\strutbox\relax}
\newcommand{\codehl}[2]{%
  \begingroup
  \setlength{\fboxsep}{0pt}%  
  \colorbox{#1}{\reducedstrut#2\/}%
  \endgroup
}

\crefname{lstlisting}{listing}{listings}
\Crefname{lstlisting}{Listing}{Listings}

\newcommand{\as}{\overline{a}}
\newcommand{\xs}{\overline{x}}

\newcommand{\ds}{\overline{d}}

\newcommand{\Subst}{\mathcal{S}}
\newcommand{\Tsubst}{\mathcal{T}}
\newcommand{\Isubst}{\mathcal{I}}
\newcommand{\Map}{\mathcal{M}}
\newcommand{\plet}{\xspace{\tt let}~}                                                
\newcommand{\pin}{~{\tt in}~}                                                       
\newcommand{\addop}{{\tt add}}
\newcommand{\matmulop}{{\tt matmul}}
\newcommand{\transposeop}{{\tt transpose}}
\newcommand{\fop}{{\tt f}}

\newcommand{\names}{names\xspace} % names
\newcommand{\name}{name\xspace}  % name 
\newcommand{\named}{named\xspace} % nameed
\newcommand{\Named}{Named\xspace} % Named
\newcommand{\dimname}{dim\_name\space} % dim\_name
\newcommand{\toast}{{\textrm TOAST}\xspace}
\newcommand{\NDA}{\texttt{NDA}\xspace}
\newcommand{\NDAv}{\ensuremath{\mathtt{NDA}_1}\xspace}
\newcommand{\NDAw}{\ensuremath{\mathtt{NDA}_2}\xspace}

\begin{document}

% \title{Supercharging automatic partitioning of machine learning models through static program analysis}
\title{\toast{}: Fast and scalable auto-partitioning based on principled static analysis}

% \begin{mlsysauthorlist}
% \mlsysauthor{Sami Alabed}{equal, gdm}
% \mlsysauthor{Dominik Grewe}{equal, gdm}
% \mlsysauthor{Norman A. Rink}{equal, gdm}
% \mlsysauthor{Masha Samsikova}{equal, gdm}
% \mlsysauthor{Timur Sitdikov}{equal, gdm}
% \mlsysauthor{Agnieszka Swietlik}{equal, gdm}
% \mlsysauthor{Dimitrios Vytiniotis}{equal, gdm}
% \mlsysauthor{Daniel Belov}{gdm}

\author[S.~Alabed]{Sami Alabed}
\authornote{Equal contribution, authors in alphabetical order. Correspondence to: \url{dvytin@google.com}}
\affiliation{\institution{Google DeepMind} \city{London} \country{UK}}
\orcid{0000-0001-8716-526X}

\author[D.~Grewe]{Dominik Grewe}
\authornotemark[1]
\affiliation{\institution{Isomorphic Labs} \city{London} \country{UK}}
\authornote{Work done while at Google DeepMind}
\orcid{0009-0008-6483-3841}

\author[N.~A.~Rink]{Norman A. Rink}
\authornotemark[1]
\affiliation{\institution{Google DeepMind} \city{London} \country{UK}}
\orcid{0009-0000-8591-5215}

\author[M.~Samsikova]{Masha Samsikova}
\authornotemark[1]
\affiliation{\institution{Google DeepMind} \city{London} \country{UK}}
\orcid{0009-0009-1125-4974}

\author[T.~Sitdikov]{Timur Sitdikov}
\authornotemark[1]
\affiliation{\institution{Google DeepMind} \city{London} \country{UK}}
\orcid{0009-0007-4010-8912}

\author[A.~Swietlik]{Agnieszka Swietlik}
\authornotemark[1]
\affiliation{\institution{Google DeepMind} \city{London} \country{UK}}
\orcid{0009-0003-7276-9038}

\author[D.~Vytiniotis]{Dimitrios Vytiniotis}
\authornotemark[1]
\affiliation{\institution{Google DeepMind} \city{London} \country{UK}}
\orcid{0009-0007-2079-1996}

\author[D.~Belov]{Daniel Belov}
\affiliation{\institution{Google DeepMind} \city{London} \country{UK}}
\orcid{0009-0003-8374-1874}

\renewcommand{\shortauthors}{Sami Alabed et al.}

% \end{mlsysauthorlist}
% \mlsysaffiliation{gdm}{Google DeepMind}
% \mlsysaffiliation{iso}{Isomorphic Labs}
% \mlsyscorrespondingauthor{Dimitrios Vytiniotis}{dvytin@google.com}

\begin{abstract}
Partitioning large machine learning models across distributed accelerator systems is a complex process, requiring a series of interdependent decisions that are further complicated by internal sharding ambiguities.
Consequently, existing auto-partitioners often suffer from out-of-memory errors or are prohibitively slow when exploring the exponentially large space of possible partitionings. 
To mitigate this, they artificially restrict the search space, but this approach frequently yields infeasible solutions that violate device memory constraints or lead to sub-optimal performance.

We propose a system that combines a novel static compiler analysis with a Monte Carlo Tree Search. 
Our analysis constructs an efficient decision space by identifying (i) tensor dimensions requiring identical sharding, and (ii) partitioning ``conflicts'' that require resolution.

Our system significantly outperforms state-of-the-art industrial methods across diverse hardware platforms and model architectures, discovering previously unknown, superior solutions, and the process is fully automated even for complex and large models.
% Notably, the discovered solutions often involve internal reshardings and, in some cases, previously unknown partitioning strategies.
\end{abstract}

\maketitle

\section{Introduction}
\label{sec:introduction}
The increasing size of machine learning (ML) models necessitates training on distributed systems of accelerator devices (e.g.,~GPUs, TPUs \cite{DBLP:journals/corr/JouppiYPPABBBBB17, jouppi+:lessons, tpu_cloud}).
This distributed approach is required because the memory footprints of model parameters and input data routinely exceed the capacity of a single device.
Consequently, ML engineers face the significant challenge of adapting their model code for execution on these distributed systems.
Manually converting code that defines the mathematics of a model to semantically equivalent code that executes on a distributed system is time-consuming, highly error prone and simply does not scale with model size and complexity.
To assist with this challenge, a number of partitioning tools \cite{gshard, gspmd2021, shazeer2018mesh, partir24_full} have emerged.
These tools typically take in ML code in a domain-specific language (e.g. TensorFlow \cite{tensorflow_osdi_2016}, JAX \cite{jax2018github}, PyTorch \cite{pytorch2019}) and, usually at the level of an intermediate representation (IR), rewrite code for local device execution, inserting cross-device communication.
This local code operates on partitioned tensors, specifically the {\em shards} of inputs and model parameters residing on each device.
While these tools automate the mechanics of partitioning a model, users are still responsible for determining the optimal sharding strategy.
This requires deciding which tensors to partition based on the model architecture, tensor dimensions, and device topology.
Developing optimal partitioning strategies requires significant engineering effort \cite{megatron2019, zero2019, fsdp, edge_sharding,seq_paralleism_nvidia, inference_transformer}.
Consequently, many new model architectures lack well-established partitioning strategies.

Automatic partitioning tools \cite{automap, alpa2022} have emerged to discover partitioning strategies for ML models.
% with little to no user guidance.
These tools operate by exploring the search space of possible model partitionings, guided by a performance model that estimates the runtime of each configuration.
This search space is exponential in the number of model operations as it comprises all possible sharding combinations for every tensor.
Even for models represented in a high-level intermediate representation (IR) like StableHLO \cite{stablehlo}, this involves tens of thousands of operations, making exhaustive exploration intractable.
For instance, Alpa \cite{alpa2022} considers every tensor as a candidate for sharding, creating a large search space, which leads to slow compilation times and yields sub-optimal solutions.
AutoMap \cite{automap, automap-reloaded} addresses this challenge by combining {\em explicit} sharding decisions, exposed in the search space, with {\em implicit} compiler-driven {\em propagation} of those decisions to the rest of the model.
However, this approach relies on the user to manually specify key locations in the model where resharding might be beneficial (e.g., for collective communication in transformer sequence parallelism~\cite{seq_paralleism_nvidia}).
Therefore, auto-partitioners must aggressively prune the search space without discarding good solutions.

% Automatic partitioning tools \cite{automap, alpa2022} emerged to automatically discover good partitionings of ML models, relying on little to no user guidance.
% This is achieved by exploring the space of possible model partitionings, where exploration is be guided by a metric that estimates their final runtime performance.
% The full space of possible partitionings combines all possible shardings of all tensors in the model.
% This space is exponential in the number of operations in the model, of which there may be 10k--100k already if the model is represented in a not-too-low-level tensor IR, e.g.~StableHLO \cite{stablehlo}.
% Hence, to have any hope of discovering useful partitionings within a practical budget, auto-partitioners must disregard a large portion of this space, without missing interesting partitionings. 

% Despite its remarkable results, Automap failed to expose sharding decisions for intermediate values in model code, which is needed for scaling the context dimension of large Transformers \cite{DBLP:journals/corr/VaswaniSPUJGKP17}).
% It also relied on continuously running the costly compiler propagation, which rewrites large model code, during search.

In this paper, we address the challenge of automatically expanding the search space to include interesting sharding strategies for intermediate tensors, without making this space intractably large. Our contributions are as follows:
% In this paper the main challenge we address is how to effectively and automatically expose a larger space that includes interesting shardings of intermediate values, without making the space too large and impractical for exploration. Specifically our contributions are: 
\begin{itemize}
    \item A novel static analysis that operates ahead of the partitioning search to identify tensor dimensions that must be sharded identically.
    % which we call the {\em colored dimensions analysis} (\NDA),
    % % \nrink{Should not stress the more-than-one uses here. We do not really get back to it.}
    % %(for example input activations in transformers)
    % have the potential to create {\em sharding conflicts}, situations where the multiple datapaths of a value end up introducing the need for different shardings.
    % \item We show how {\em sharding conflicts} may occur, a situation where it is genuinely ambiguous how to shard an operation based on the shardings of its operands.
    % By maintaining a distinction between definitions and uses of tensors, our analysis can pinpoint conflicts.
    % We show with examples, and validate through our evaluation, how the interesting space of internal reshardings corresponds to the number of different
    % {\em orders} of sharding the dimensions participating in a conflict. (\Cref{sec:NDA:conflicts})
    
    \item A method for identifying and leveraging {\em sharding conflicts}—situations where an operation's sharding is ambiguous based on its operands. 
    We demonstrate that these conflicts precisely define the search space for beneficial internal resharding strategies (\Cref{sec:NDA:conflicts}).

    % \nrink{We do not really show that. It comes out post-hoc in our evaluation of scaling transformers.}
    % \dvytin{need section refs}
    %
   
    % \item A new automatic partitioning tool by co-designing a Monte-Carlo Tree Search agent~\cite{Browne12asurvey} with our static analysis, which augments the search space with {\em conflict resolution actions} and prune the search space by identifying dimensions that are sharded together, enabling it to discover complex resharding strategies efficiently (\Cref{sec:toast}).
    \item \textbf{T}he \textbf{o}ther \textbf{a}uto-\textbf{s}harding \textbf{t}ool (\toast{}).
An automatic partitioning tool that integrates our analysis with a Monte-Carlo Tree Search (MCTS) agent~\cite{Browne12asurvey}.
The analysis guides the agent by (1) augmenting the search space with {\em conflict resolution actions} and (2) pruning the space by identifying dimensions that must be sharded together.
This design enables the efficient discovery of complex resharding strategies (\Cref{sec:toast}).
    % \item 
    %  We co-designed our static analysis and Monte-Carlo Tree Search~\cite{Browne12asurvey} agent into a new automatic partitioning tool: Auto sharding with colors (\toast{}, \Cref{sec:toast}). \toast{} takes explicit sharding actions on named dimensions to also take into account {\em conflict resolution orders} for conflicting dimensions, directly generated by our static analysis in the previous step. Our MCTS agent is driven, as in previous
    % work, by a simulated runtime cost model subject to memory constraints. 
    % In addition, our MCTS
    % agent includes the following advancements over previous work: (i) further action pruning based on caching the effects of available actions on the MCTS state (Section~\dvytin{XX}), (iii) argument grouping for parameters of repeated layers in a model, which also uses the NDA for more robust semantic pattern matching. 
    
    \item A comprehensive evaluation (\Cref{sec:eval}) demonstrating that \toast{} consistently outperforms state-of-the-art automated and expert partitioning strategies across various ML models and hardware platforms. 
    % Our results also show that \toast{} enables scaling models to sizes unattainable by existing methods.
\end{itemize}

\section{Background}
\label{sec:back}

\subsection{Partitioning of ML models}
\label{sec:back:model-partitioning}
% In the context of ML models, it is a common abstraction to organize the available accelerator devices (e.g. GPUs, TPUs) into a (logical) {\em mesh}, i.e.~into an $n$-dimensional lattice that is spanned by $n$ {\em device axes} (cf.~\cite{gshard, gspmd2021, shazeer2018mesh, partir24_full}).

When scaling ML models, it is a common abstraction to organize available accelerator devices (e.g., GPUs, TPUs) into a logical {\em mesh}.
This mesh is an $n$-dimensional lattice spanned by $n$ {\em device axes}~\cite{gshard, gspmd2021, shazeer2018mesh, partir24_full}.
In this section, we assume that devices are organized in a 2-dim.~mesh, spanned by axes \texttt{b} and \texttt{m}.
If there are $b$ devices along axis \texttt{b} and $m$ devices along \texttt{m}, then the whole mesh consists of $b\cdot m$ devices.

% {\em Partitioning} an ML model means preparing it for execution on a mesh of devices. For partitioning to be effective, one wants the devices in the mesh to operate on {\em shards} of tensors.
% Shards are obtained by slicing a tensor along one of its dimensions.
% These shards are then distributed across the devices along one of the axes of the mesh, as in \Cref{fig:sharded-tensor}.

\begin{figure}
    \centering
    \includegraphics[scale=0.26]{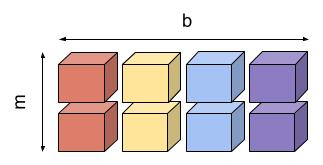}
    \hspace{6pt}
    % \includegraphics[scale=0.26]{figures/sharding m only.jpg}
    % \hspace{6pt}
    \includegraphics[scale=0.26]{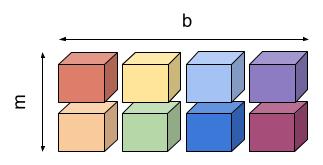}
    \caption{%
        Shards of a 3-dim.~tensor on a 2-dim.~mesh (axes \texttt{b} and \texttt{m}).
        Blocks correspond to different devices, colors correspond to different shards (i.e.~different data).
        Left: sharding along axis \texttt{b}.
        Right: sharding along axes \texttt{b} and \texttt{m}.
    }
    \label{fig:sharded-tensor}
\end{figure}

{\em Partitioning} a ML model is the process of preparing it for execution on a mesh of devices.
It involves distributing the model's tensors across the devices as smaller chunks, known as {\em shards}.
A shard is created by slicing a tensor along one of its dimensions and assigning it to devices along a corresponding axis of the device mesh, as illustrated in \Cref{fig:sharded-tensor}.

% A tensor can be sharded along multiple of its dimensions, as indicated in the right pane of \Cref{fig:sharded-tensor}.

\begin{figure*}
\begin{minipage}[t]{0.31\linewidth}
\begin{lstlisting}[
    language=python,
    basicstyle=\ttfamily\scriptsize,
    label={lst:linear-layers:highlights}
]
def mlp(x  : [(*@\codehl{yellow}{256}@*), 32],
        w1 : [32, (*@\codehl{green}{64}@*)],
        w2 : [(*@\codehl{green}{64}@*), 16]) {
  y  : [(*@\codehl{yellow}{256}@*), (*@\codehl{green}{64}@*)] = matmul(x, w1)
  z  : [(*@\codehl{yellow}{256}@*), (*@\codehl{green}{64}@*)] = ReLU(y)
  w  : [(*@\codehl{yellow}{256}@*), 16] = matmul(z, w2)
  return w
}
\end{lstlisting}%
%\vspace{8pt}
\subcaption{Simple ML model: two linear layers.}
\label{fig:linear-layers:highlights}
\end{minipage}
\begin{minipage}[t]{0.32\linewidth}
\begin{lstlisting}[
    language=python,
    basicstyle=\ttfamily\scriptsize,
    label={lst:linear-layers:batch}
]
def mlp(x  : [256{b}, 32],
        w1 : [32, (*@\codehl{green}{64}@*)],
        w2 : [(*@\codehl{green}{64}@*), 16]) {
  y  : [256{b}, (*@\codehl{green}{64}@*)] = matmul(x, w1)
  z  : [256{b}, (*@\codehl{green}{64}@*)] = ReLU(y)
  w  : [256{b}, 16] = matmul(z, w2)
  return w
}
\end{lstlisting}%
%\vspace{8pt}
\subcaption{Batch partitioning along device axis \texttt{\{b\}}.}
\label{fig:linear-layers:batch}
\end{minipage}
\begin{minipage}[t]{0.34\linewidth}
\begin{lstlisting}[
    language=python,
    basicstyle=\ttfamily\scriptsize,
    label={lst:linear-layers:batch-and-model}
]
def mlp(x  : [256{b}, 32],
        w1 : [32, 64{m})],
        w2 : [(64{m}, 16]) {
  y  : [256{b}, 64{m}] = matmul(x, w1)
  z  : [256{b}, 64{m}] = ReLU(y)
  w_ : [256{b}, 16] = matmul(z, w2)
  w  : [256{b}, 16] = all_reduce {m} w_
  return w
}
\end{lstlisting}%
\subcaption{Batch and model partitioning.}
\label{fig:linear-layers:batch-and-model}
\end{minipage}
\caption{Two-layer MLP with annotations. We highlight 
dimensions that should be sharded the same way.}
\label{fig:linear-layers}
\end{figure*}%

% The simplified {\em multi-layer perceptron} (\texttt{mlp}) in \Cref{fig:linear-layers:highlights} illustrates these concepts.
% There are two matrix multiplications, and a non-linear operation (\texttt{ReLU}) in between.
% The inputs, or {\em samples}, are passed in as \texttt{x},
% and the model parameters are given as arguments \texttt{w1}, \texttt{w2}.
% We observe that all operations in the body of \texttt{mlp} act as maps across the tensor dimensions highlighted in yellow: \colorbox{yellow}{\tt 256}.
% Thus, all operations can be executed in parallel on shards of tensors along these dimensions.
The simplified {\em multi-layer perceptron} (\texttt{mlp}) in \Cref{fig:linear-layers:highlights} illustrates these concepts.
The model consists of two matrix multiplications separated by a non-linear operation (\texttt{ReLU}).
It takes the input samples, \texttt{x}, and model parameters, \texttt{w1} and \texttt{w2}, as arguments.
All operations in the body of \texttt{mlp} are mapped across the tensor dimension highlighted in yellow: \colorbox{yellow}{\tt 256}.
Consequently, these operations can be executed in parallel on tensor shards created along this dimension.
\Cref{fig:linear-layers:batch} uses the annotation \texttt{\{b\}} to indicate that these dimensions have been sharded along axis \texttt{b}.
Although the body of \texttt{mlp} appears unchanged, its operations now execute on smaller tensors, specifically the shards corresponding to the full tensors shown in \Cref{fig:linear-layers:highlights}.
% 
% \Cref{fig:linear-layers:batch} is the {\em batch partitioned} version of the original \texttt{mlp} function:
% the function now operates on only a shard of \texttt{x}, 
% and a shard of the samples \texttt{x} is commonly referred to as a {\em batch}.
% Note that \Cref{fig:linear-layers:batch} is the {\em device-local} code for \texttt{mlp}.
% Every device initially holds a shard of \texttt{x}.
% During execution of the code in \Cref{fig:linear-layers:batch}, each device computes shards of intermediate tensors, and the returned result \texttt{w} on each device represents a shard of the the full tensor in \Cref{fig:linear-layers:highlights}. 
%
\Cref{fig:linear-layers:batch} is the {\em batch-partitioned} version of the \texttt{mlp} function. 
The function now operates on a shard of the input samples \texttt{x}, known as a {\em batch}. 
This listing shows the {\em device-local} code for \texttt{mlp}, where each device initially holds one such shard of \texttt{x}. 
During execution, each device computes shards of intermediate tensors. Consequently, the returned tensor \texttt{w} on each device represents a shard of the full output tensor from \Cref{fig:linear-layers:highlights}.
Our batch partitioning of \texttt{mlp} requires no communication between devices.
% The shards of \texttt{w} on different devices must be concatenated along the leading dimension to interpret them as the global result \texttt{w} that is computed in \Cref{fig:linear-layers:highlights}.%
% \footnote{%
% When training real ML models, this concatenation is usually avoided:
% sharded results computed in one training step are fed back into the partitioned model as sharded tensors in the next training step.
% }

%\footnote{Batch partitioning does require communication  back-propagation.}
Generally, partitioning relies on cross-device communication to ensure that the distributed execution of a model yields results identical to the results computed by the unpartitioned model.
MPI-style primitives are typically used for collective communication between devices in the mesh \cite{MPI:1995}. 
Next, we develop an example of this.
The \texttt{mlp} function in \Cref{fig:linear-layers:batch} can be further partitioned.
For example, partitioning the yellow dimension (\colorbox{yellow}{\tt 256}) along additional device axes, leading to smaller batches being processed by each device.
But note how the parameters {\tt w1} and {\tt w2} did not get partitioned this way, even though we may want to to shard them to fit memory constraints. 
% However, the parameters {\tt w1} and {\tt w2} are not partitioned in this manner and may require sharding to satisfy memory constraints.
To achieve this, 
observe that the first \texttt{matmul} and the \texttt{ReLU} operations act as maps across the dimensions highlighted in green (\colorbox{green}{\tt 64}). Hence, sharding these dimensions again leads to independent device-local computations, up to the definition of \texttt{z}.
In the final \texttt{matmul}, however, the green dimension is reduced over.
If each device operates on a shard of \texttt{z} (and a shard of \texttt{w2}), only a contribution to this reduction is computed locally.
To sum up the contributions from the devices, \texttt{all\_reduce}-style communication is required  after computing the device-local \texttt{matmul}s.

\Cref{fig:linear-layers:batch-and-model} shows the device-local code for \texttt{mlp} after sharding the green dimensions along axis \texttt{m} (after batch partitioning along \texttt{b}).
The final reduction is performed across axis \texttt{m}, indicated by the attribute \texttt{\{m\}} on the \texttt{all\_reduce}.
This kind of partitioning model parameters that are used in pairs of \texttt{matmul}s is the essence of Megatron partitioning \cite{megatron2019}.

\subsection{Partitioning, algorithmically}
\label{sec:back:algorithmic}
% These dimensions need not always appear in the same position:
% in \Cref{fig:linear-layers:batch}, the green dimension is the leading one in \texttt{w2} but the trailing one in \texttt{w1}.

% We have have exhibited the well-known batch and Megatron partitioning on our {\tt mlp} example.
% For each strategy we had to shard the right set of tensors, each along a specific dimension, i.e.~the colored dimensions in \Cref{fig:linear-layers:highlights}.
% We found the correct dimensions to shard simply by inspecting the operations in \texttt{mlp}, trying to
% minimize communication.

% In this manual approach, the correct dimensions were identified by inspecting the operations in the \texttt{mlp} with the goal of minimizing communication.

We demonstrated batch and Megatron partitioning strategies on our \texttt{mlp} example.
Each strategy requires sharding a specific set of tensors along particular dimensions, corresponding to the colored dimensions in \Cref{fig:linear-layers:highlights}.
In manual approach, the dimensions were identified by inspecting the operations in the \texttt{mlp} to minimize communication.

How can a tool discover the right sets of dimensions? Simply enumerating all possible partitionings for all tensors does not scale with the ML model size.
% Assume that the body of a function that is to be partitioned has $k$ operations.
% If an operation defines one tensor with $d$ dimensions, we then estimate that there are about $(d+1)$ ways of partitioning that operation, along a single device axis.
% This is because the available device axis can be used to partition one of the $d$ dimensions, or none at all.
% This gives a total of about $(d+1)^k$ partitionings of the whole function, demonstrating the exponentially large size of the space of all possible partitionings.
Hence the need for a systematic method to focus on candidates for partitionings.

\paragraph{How it was done before.}
Tools like Automap~\cite{automap, automap-reloaded}, rely on compiler-based {\em propagation} of tensor shardings found in partitioning engines~\cite{partir24_full, gspmd2021, shardy}.
E.g.~in the \texttt{mlp} function above, a search agent would {\em only} issue a sharding of the 64-sized (green) dimension of \texttt{w1}.
The compiler would propagate this sharding to partition the first \texttt{matmul}, then through the \texttt{ReLU}, reach the second \texttt{matmul}, and finally 
propagate the sharding to the other parameter \texttt{w2}, and emit an \texttt{all-reduce}.
% Compiler propagation is a complex combination of {\em forward}, {\em backward}; and
% {\em operand-to-operand} analysis -- following standard practice in other partitioning engines like GSPMD~\cite{gspmd2021}.
Alpa~\cite{alpa2022} encodes the sharding as a logical mapping of dimensions to sets of devices, and employs internal propagation-like heuristics to simplify these constraints.
% before sending them off to constraint solvers.

\paragraph{Enter \named dimensions.}
If we had a way to identify, ahead of partitioning time, the dimensions that should be sharded together -- i.e.~dimensions with the same color in \Cref{fig:linear-layers:highlights} -- then partitioning becomes simpler: 
pick an axis \texttt{a} and a color \texttt{C}, and shard all tensors whose dimensions include \texttt{C} along the dimension colored with \texttt{C}.
% \footnote{Unless the tensor is already sharded across axis \texttt{a}, a coner care.}
This is the idea we build on for the rest of the paper.

The static analysis we present in \Cref{sec:NDA} discovers sets of dimensions analogous to the colored dimensions in \Cref{fig:linear-layers:highlights}
by carefully identifying {\em logical dimension \names} based on rules that describe the parallel and contracting dimensions for every operation. % in our underlying IR. 
% This results 
% To achieve this, observe that partitioning a whole function amounts to partitioning (subsets of) the operations in the body of the function.
% Then, the colored sets of dimensions in \Cref{fig:linear-layers:highlights} are obtained by partitioning operations along a dataflow path in ways that fit together, in the sense that no communication should be required to feed the sharded tensor produced by one operation into the partitioned operation that consumes the tensor.
These sets of dimensions act as the search space for our automatic partitioner (\Cref{sec:toast}).
% Once we have these sets of dimensions, we explore the space of possible partitionings with a Monte-Carlo Tree Search (MCTS) (\Cref{sec:toast:MCTS}) that successively shards tensors along named (viz.~colored) dimensions, one axis at a time.

\paragraph{\Named dimensions for resolving sharding conflicts.}
Beyond the apparent simplicity, our analysis is able to identify, ahead of time, situations where different dataflow paths originating from the same tensor lead to sharding ambiguities.
Function \texttt{f} below provides an example:
\texttt{x} flows into the \texttt{matmul} along two dataflow paths, directly and through \texttt{y}.
\begin{lstlisting}[
    language=python,
    basicstyle=\ttfamily\small
]
def f(x : [32, 4]) {
  y : [4, 32] = transpose(x)
  z : [32, 32] = matmul(x, y)
  return z
}
\end{lstlisting}
Propagating the sharding of \texttt{x} along both paths leads to an ambiguity for sharding the \texttt{matmul}. For
example, if \texttt{x} is sharded on its first dimension (of size 32), then \texttt{y} is sharded on
the second dimension, and it is now unclear whether \texttt{matmul} should become sharded along its first or second dimension.
Conceptually, if we were to assign colors to dimensions, \texttt{matmul} would receive {\em 
the same color on both dimensions}, a situation that we call a {\em sharding conflict}.
When partitioning the \texttt{matmul} operation along that color, we need to pick which one of the two dimensions of that color to shard. 

Our analysis is not only able to discover these ambiguities but -- going further -- we show that by keeping track of {\em different
dimension \names for different uses of tensors} we are able to extract all the possible choices for resolving these
sharding conflicts (\Cref{sec:NDA:conflicts}), and later surface them as part of the search space (in MCTS).
This ability is key to enabling \toast{} to partition attention computations~\cite{DBLP:journals/corr/VaswaniSPUJGKP17}, the cornerstone of modern Large Language Models, at context lengths beyond what other methods are capable of.

% As our evaluation shows
% Our experimental evaluation justified post-hoc that we do indeed discover good partitionings based on the dimension sets from our static analysis, and within a reasonable search budget.

%\input{sections/arch}

\begin{figure*}
\footnotesize
\begin{minipage}[t]{0.33\linewidth}
%\vspace{0pt}
\begin{equation*}
\begin{array}{lcll}
    \multicolumn{3}{l}{\textbf{Tensor expressions in ANF}} \\
    e &  ::= & x                         & \text{variable} \\
      & \mid & \plet x = op(\xs) \pin e  & \text{local definition} \\[1mm]
    \multicolumn{4}{l}{
        op \in \{\fop, \addop, \transposeop, \matmulop, \ldots\}}
\end{array}
\end{equation*}
\end{minipage}
\begin{minipage}[t]{0.33\linewidth}
% %\vspace{0pt}
\begin{equation*}
\begin{array}{lcll}
    \multicolumn{4}{l}{\textbf{Auxiliary definitions}} \\
    \multicolumn{3}{l}{a,b,c,d,d_1, \dots, d_k}     & \text{dimension \names} \\
    \ds     & ::= & [] \mid [d_1, \dots, d_k]      & \text{\Named dimensions} \\
    E       & ::= & \cdot \mid E, x : \ds     & \text{Environment}
\end{array}%
\end{equation*}
\end{minipage}%
\begin{minipage}[t]{0.30\linewidth}
% %\vspace{0pt}
\begin{equation*}
\begin{array}{lcll}
    \multicolumn{4}{l}{} \\
    \Map    & ::= & \emptyset \mid \Map \cup \{c\mapsto d\}  & \text{Map} \\
    \Isubst  & ::= & \emptyset \mid \Isubst \cup \{d\circeq c\} & \text{Identities}
\end{array}
\end{equation*}
\end{minipage}%
\\[2mm]
\begin{minipage}[t]{\linewidth}
\begin{equation*}
\begin{array}{l}
    \textbf{\Named Dimensions Analysis} \\
    \NDA_{E:\text{Environment}} : (e:\text{Expression}) \rightarrow (\ds : \text{\Named dimensions}) \times (\Map:\text{Map}) \times (\Isubst:\text{Identities}) \\ \\ 
\end{array}%
\end{equation*}
\\
\begin{minipage}[t]{0.58\linewidth}
% %\vspace{0pt}
\begin{equation*}
\begin{array}{lcll}
\NDA_{E}(\plet x = op(\xs) \pin e) 
& := &
    (\ds_2, \Map_1\cup\Map_2, \Isubst_1\cup\Isubst_2), & \\
\textsc{(let)}
& & \text{where}\: (\ds_1, \Map_1, \Isubst_1) = \NDA_{E}(op(\xs)), \\
& & \phantom{\text{where}}\: (\ds_2, \Map_2, \Isubst_2) = \NDA_{E, x:\ds_1}\!\!(e)
\\[2mm]
\NDA_{E}(\mathtt{reduce}_{r,op}(x)) & := &
    ([a_1, \ldots, a_{r-1}, a_{r+1}, \ldots, a_k], \Map, \{a_i\circeq d_i\}), & \\
\textsc{(reduce)}
& & \text{where}\: ([d_1, \ldots, d_r, \ldots, d_k], \Map, \emptyset) = \NDA_{E}(x), & \\
& & \text{with $a_i$ \textit{fresh}},\,
    \text{for}\: op\in\{\mathtt{add},\mathtt{mul},\ldots\} &
\\[2mm]
\NDA_{E}(\mathtt{matmul}(x, y)) & := & 
    ([a_1, a_2], \Map_1\cup\Map_2, \{a_1 \circeq d_1, a_2 \circeq c_2, d_2 \circeq c_1\}), & \\
\textsc{(matmul)}
& & \text{where}\: ([d_1, d_2], \Map_1, \emptyset) = \NDA_{E}(x), \\
& & \phantom{\text{where}}\: ([c_1, c_2], \Map_2, \emptyset) = \NDA_{E}(y),\,
    \text{with}\: a_1,a_2\:\textit{fresh}
\end{array}
\end{equation*}
\end{minipage}
\begin{minipage}[t]{0.40\linewidth}
% %\vspace{0pt}
\begin{equation*}
\begin{array}{lcll}
\NDA_{E}(x) & := &
    (\as, \{d_i\mapsto a_i\}, \emptyset), &  \\
\textsc{(variable use)}
& & \text{if}\: x : \ds \in E,\,
    \text{with $a_i$ \textit{fresh}}
\\[2mm]
\NDA_{E}(\mathtt{f}(x)) & := &
    (\as, \Map, \{a_i\circeq d_i\}), &  \\
\textsc{(function)}
& & \text{where}\: (\ds, \Map, \emptyset) = \NDA_{E}(x), \\
& & \text{with $a_i$ \textit{fresh}}
\\[2mm]
\NDA_{E}(op(x, y)) & := &
    (\as, \Map_1\cup\Map_2, \{a_i\circeq d_i, a_i\circeq c_i\}) & \\
\textsc{(op)}
& & \text{where}\: ([d_1, \ldots, d_k], \Map_1, \emptyset) = \NDA_{E}(x), \\
& & \phantom{\text{where}}\: ([c_1, \ldots, c_k], \Map_2, \emptyset) = \NDA_{E}(y), \\
& & \text{with}\: a_i\:\textit{fresh},\,
    \text{for}\: op\in\{\mathtt{add},\mathtt{mul},\ldots\}
\end{array}
\end{equation*}
\end{minipage}
\\[1mm]
% \begin{minipage}[t]{\linewidth}
\begin{minipage}[t]{0.58\linewidth}
\begin{equation*}
\begin{array}{lcll}
\NDA_{E}(\mathtt{transpose}_{lr}(x)) & := &
    ([a_1, \ldots, a_r, \ldots, a_l, \ldots, a_k], \Map, \{a_i\circeq d_i\}),  & \\
\makebox[\widthof{$\NDA_{E}(\plet x = op(\xs) \pin e)$}][l]{\textsc{(transpose)}}
& & \text{where}\: ([d_1, \ldots, d_l, \ldots, d_r, \ldots, d_k], \Map, \emptyset) = \NDA_{E}(x),\,
    \text{with}\: a_i\:\textit{fresh}
\\[2mm]
\NDA_{E}(\mathtt{broadcast}_{i}(x)) & := &
    ([a_1, \ldots, a_{l-1}, a, a_l, \ldots, a_k],
     \Map,
     \{a_1\circeq d_1,\ldots, a_{l-1}\circeq d_{l-1}, a_l\circeq d_l,\ldots, a_k\circeq d_k\}) & \\
\makebox[\widthof{$\NDA_{E}(\plet x = op(\xs) \pin e)$}][l]{\textsc{(broadcast)}}
& & \text{where}\: ([d_1, \ldots, d_{l-1}, d_l, \ldots, d_k], \emptyset, \Map) = \NDA_{E}(x),\,
    \text{with}\: a_i,\ldots,a_{l-1},a,a_l,\dots,a_k\:\textit{fresh}
\end{array}
\end{equation*}
\end{minipage}
    \caption{The \named dimensions analysis (\NDA), illustrating the cases for several ops}
    \label{fig:NDA}
\end{minipage}
\end{figure*}

\section{\Named Dimension Analysis}
\label{sec:NDA}
%(samialab): I think you want to start with a motivation behind this a bit, something higher level before you jump on how the NDA is assigning unique fresh name.
% why is it done, what is the goal of it etc...
We now describe our static analysis that discovers sets of dimensions analogous to the colored dimensions in \Cref{fig:linear-layers:highlights}.
Each op in the program is assigned unique dimensions for its operands and its results, some of which we identify (unify) by inspecting a rule that tells
us how the dimensions of operands or results can be sharded together. For example, a {\tt matmul} can be computed in a sharded fashion along
its first dimension if provided with shards of its first operand also on the first dimension. The rules are populated ahead of time for every op
in our array IR, analogously to~\cite{partir24_full, shardy}.
% Every operation is assigned unique dimensions for its operands and its results. 
% For each operation in a program, we then inspect a rule that tells us how the dimensions of tensors that participate in this operation should be identified for
% sharding purposes.

% \dvytin{I think here we could say a few more words about linear algebra homomorphisms. What does it mean to
% "identify" dimensions, really? (more about this in the matmul example below)}
% \nrink{
% % The rules capture how an operation produces a shard of its result from shards of its operand tensors:
% % The dimensions along which operands and results must be sharded for this are the ones that are identified by our rules.
% we identify dimensions of operand and result tensors precisely if all these dimensions must be sharded for the operation to compute a shard of its result from shards of its operands.
% }
After identifying dimension \names based on all the rules that apply to the operations in a program, each remaining (unidentified) dimension \name will appear in a set of dimensions analogous to the colored dimensions in \Cref{fig:linear-layers:highlights}.

\subsection{Definition of the analysis}
\label{sec:NDA:definition}
\Cref{fig:NDA} defines our {\em \named dimensions analysis} ($\NDA$) as a recursive function that operates on straight line tensor programs in A-normal form \cite{Sabry:1992}, which
is equivalent to the SSA form used in our implementation. 
Given an environment $E$ that maps free variables to dimension \names, the $\NDA$ computes a triple consisting of:
\begin{enumerate}[(i)]
    \item \label{item:assigns-named-dimensions}
   an assignment of dimension \names to tensor op results,
    \item \label{item:computes-map}
    a mapping $\Map$ connecting dimension \names of value definitions to the dimension \names of their uses,
    %(from variable definitions to their uses)
    and
    \item \label{item:computes-substitution}
    a set of identities $\Isubst$ between dimension \names.
\end{enumerate}
The dimension \names assigned in \ref{item:assigns-named-dimensions} are the $a_i$ that appear in \Cref{fig:NDA}.
%These are always chosen to be fresh, i.e.~none of the $a_i$ ever appear in the dimension \names of more than one operation or variable.
%The mapping $\Map$ from \ref{item:computes-map} connects dimension \names that are assigned to variable definitions, as recorded in the environment $E$, to the corresponding \names that are assigned to variable uses.
%
To illustrate \ref{item:assigns-named-dimensions} and \ref{item:computes-map}, consider the following definition and use of variable $y$, assuming $x$ is assigned dimension \names $[d_1, d_2]$ in the current environment $E$.
\begin{align*}
    \plet y = x \pin y
\end{align*}
The \textsc{let} rule in \Cref{fig:NDA} invokes the \textsc{variable use} rule to determine that the use of $x$ should be assigned fresh \names $[a_1, a_2]$ and the map $\Map$ should be populated with $\{d_1\mapsto a_1, d_2\mapsto a_2\}$.
The \textsc{let} rule then adds the assignment $y :[a_1, a_2]$ to $E$,
before invoking the \textsc{variable use} rule again, this time for the use of $y$ after \texttt{in}.
The use of $y$ is assigned fresh \names $[b_1, b_2]$, and $\Map$ is extended with $\{a_1\mapsto b_1, a_2\mapsto b_2\}$.
% Variables are defined by \plet expressions, and the case \textsc{let} in \Cref{fig:NDA} recursively calls \NDA to work out that the variable $x$ should be assigned named dimensions $\ds_1$.
% This assignment, $x:\ds_1$, is then added to the environment $E$ and is subsequently available when \NDA processes subexpression $e$ of the \plet expression.
% Inside $e$, the variable $x$ may be used, and then case \textsc{variable use} from \Cref{fig:NDA} applies:
% the assignment $x:\ds_1$ is looked up in the environment, but the use of $x$ (as an expression) is assigned fresh dimension \names $\as$.
Notice how $\Map$ connects dimension \names between definitions and uses:
$b_1$ and $b_2$ from the use of $y$ can be traced all the way back to $d_1$ and $d_2$ from the definition of $x$. 
% We stress that only the \textsc{variable use} rule in \Cref{fig:NDA} adds mappings to $\Map$;
% other rules simply thread through maps $\Map,\Map_1,\Map_2$ that stem from evaluating \NDA on subexpressions.

The identities $\Isubst$ from \ref{item:computes-substitution} record which dimensions that appear in an operation should be sharded identically.
We consider \texttt{matmul(x, y)} to illustrate this.
Assuming that the the \textsc{matmul} rule in \Cref{fig:NDA} has assigned dimension \names $[d_1,d_2]$ and $[c_1, c_2]$ to the use of $x$ and $y$, respectively,
we compactly express this instance of the \textsc{matmul} rule as
\begin{align*}
    \texttt{matmul}(x : [d_1, d_2], y : [c_1, c_2]) : [a_1, a_2],
\end{align*}
together with identities
\begin{align*}
    a_1\circeq d_1, \quad a_2\circeq c_2, \quad d_2\circeq c_1.
\end{align*}
The first identity expresses that {\tt matmul} acts as a map on the leading dimension of the first operand, which means that if the first operand
is sharded on that dimension, we can compute the {\tt matmul} in a sharded way, and concatenate the shards on the leading dimension of the result to 
obtain the original result.
The second identity is analogous.
The third one expresses that we can shard the {\tt matmul} by sharding both operands
along the contracting dimension.
(Lowering must then introduce an {\tt all\_reduce}, as in \Cref{fig:linear-layers:batch-and-model}.)

\subsection{Partitioning functions with the \NDA}
\label{sec:NDA:function-partiitoning}
\Cref{fig:linear-layers:NDA-results} shows the \texttt{mlp} function from \Cref{fig:linear-layers:highlights} annotated with the results computed by the \NDA.
Tensor variables, and their uses, are annotated with \named dimensions (not with actual shapes).
The map $\Map$ and the identities $\Isubst$ computed by the \NDA appear in comments.

In going to \Cref{fig:linear-layers:NDA-names-identified}, we have applied the identities from $\Isubst$.
This reveals the different ways in which each operation in the function body can be partitioned, in isolation, as explained in \Cref{sec:NDA:definition}.
In comments, we still display the map $\Map$, but the identities from $\Isubst$ have also been applied to $\Map$.
Note that the entries of $\Map$ still point from dimension \names in the definitions of variables (including function arguments) to the dimension \names in variable uses.

\begin{figure*}
\begin{minipage}[t]{0.48\linewidth}
\begin{lstlisting}[
    language=python,
    basicstyle=\ttfamily\footnotesize,
    label={lst:linear-layers:NDA-results}
]
def mlp(x : [B, X], w1 : [T, U], w2 : [V, W]) {
  y  : [A1, A2] = matmul(x  : [B1, X1],
                         w1 : [T1, U1])
     # from use of x:  B -> B1, X -> X1
     # from use of w1: T -> T1, U -> U1
     # from matmul: A1 $\circeq$ B1, A2 $\circeq$ U1, X1 $\circeq$ T1
  z  : [C1, C2] = ReLU(y : [D1, D2])
     # from use of y:  A1 -> D1, A2 -> D2
     # from ReLU: C1 $\circeq$ D1,  C2 $\circeq$ D2 
  w  : [E1, E2] = matmul(z  : [F1, F2],
                         w2 : [V1, W1]) 
     # from use of z:  C1 -> F1, C2 -> F2
     # from use of w2: V -> V1, W -> W1
     # from matmul: E1 $\circeq$ F1, E2 $\circeq$ W1, F2 $\circeq$ V1
  return w
}
\end{lstlisting}
% %\vspace{17pt}
\subcaption{\NDA results, entries of $\Map$ and $\Isubst$ in comments}
\label{fig:linear-layers:NDA-results}
\end{minipage}
\begin{minipage}[t]{0.48\linewidth}
\begin{minipage}[t]{\linewidth}
\begin{lstlisting}[
    language=python,
    basicstyle=\ttfamily\footnotesize,
    label={lst:linear-layers:NDA-names-identified}
]
def mlp(x : [B, X], w1 : [T, U], w2 : [V, W]) {
  y  : [A1, A2] = matmul(x  : [A1, X1],
                         w1 : [X1, A2])
     # B -> A1, X -> X1, T -> X1, U -> A2
  z  : [C1, C2] = ReLU(y : [C1, C2])
     # A1 -> C1, A2 -> C2
  w  : [E1, E2] = matmul(z  : [E1, V1],
                         w2 : [V1, E2]) 
     # C1 -> E1, C2 -> V1, V -> V1, W -> E2
  return w
}
\end{lstlisting}
% %\vspace{-14pt}
\subcaption{Dimension \names identified with $\Isubst$ only}
\label{fig:linear-layers:NDA-names-identified}
\end{minipage}
\begin{minipage}{\linewidth}
\end{minipage}
\begin{lstlisting}[
    language=python,
    basicstyle=\ttfamily\footnotesize,
    label={lst:2-layer-MLP-NDA-1-unified}
]
def mlp(x : [B, X], w1 : [X, U], w2 : [U, W]) {
  y  : [B, U] = matmul(x : [B, X], w1 : [X, U])
  z  : [B, U] = ReLU(y : [B, U])
  w  : [B, W] = matmul(z : [B, U], w2 : [U, W])
  return w
}
\end{lstlisting}
%\vspace{-14pt}
\subcaption{Dimension \names identified with $\Isubst$ and $\Map$}
\label{fig:linear-layers:NDA-fully-identified}
\end{minipage}
\caption{Application of the \NDA to the two-layer MLP from \Cref{fig:linear-layers}}
\label{fig:linear-layers:NDA}
\end{figure*}

To connect the partitionings of individual operations in a function body, 
we consider the entries of $\Map$ as identities,
%(since here each definition is mapped to just a single use color), 
and apply them to further identify dimension \names.
This leads to \Cref{fig:linear-layers:NDA-fully-identified}.
Where \texttt{B} corresponds to the yellow, and \texttt{U} to the green colors of \Cref{fig:linear-layers:highlights}.

In conclusion, identifying the dimension \names (assigned by the \NDA) with identities from $\Isubst$ {\em and} $\Map$ (also computed by the \NDA) produces sets of dimensions like the ones we found manually in \Cref{sec:back:model-partitioning}.
Sharding these sets of dimensions along device axes leads to partitionings of whole functions.

\subsection{Sharding conflicts}
\label{sec:NDA:conflicts}
While identifying dimension \names with both $\Isubst$ and $\Map$ leads to the desired sets of dimensions for sharding, it may also introduce a problematic ambiguity:
the same dimension \name may appear more than once among the \names that annotate the same variable, as for tensor $z$ below.
%(annotations as computed by the \NDA and after identifying with $\Isubst$ and $\Map$)
\begin{lstlisting}[
    language=python,
    basicstyle=\ttfamily\small
]
def f(x: [S, T]) {
  y : [T, S] = transpose(x: [S, T])
  z : [S, S] = matmul(x: [S, T], y: [T, S])
  return z
}
\end{lstlisting}
When attempting to shard all dimensions that are labeled \texttt{S}, which of the dimensions of $z$ should be sharded?
One axis cannot shard more than one dimension of a single tensor.

\begin{figure*}
\begin{minipage}[t]{0.4\linewidth}
%\vspace{0pt}
\begin{minipage}[t]{\linewidth}
%\vspace{0pt}
\begin{lstlisting}[
    language=python,
    basicstyle=\ttfamily\footnotesize,
    label={lst:simplified-attention-layer-NDA}
]
def attn(x  : [S, D] , wq : [D, H1],
         wk : [D, H1], wv : [D, H2]) {  
  k : [S, H1] = matmul(x, wk)  # keys
  v : [S, H2] = matmul(x, wv)  # values
  q : [S, H1] = matmul(x, wq)  # queries
  qt: [H1, S] = transpose(q)
  a : [S, S]  = matmul(k, qt)
  # begin: mock softmax computation (averaging)
  b : [S]     = reduce(*@$_{1,\mathtt{add}}$@*)(a)
  c : [S, S]  = broadcast(*@$_0$@*)(b)
  d : [S, S]  = div(a, c)
  # end: mock softmax computation (averaging)
  z : [S, H2] = matmul(d, v)
  return z
}
\end{lstlisting}
%\vspace{-12pt}
\subcaption{\Named dimensions, identified with $\Isubst$ and $\Map$}
\label{fig:simplified-attention-layer-NDA}
\end{minipage}
\begin{minipage}[t]{\linewidth}
%\vspace{0pt}
\begin{lstlisting}[
    language=python,
    basicstyle=\ttfamily\footnotesize,
    label={lst:simplified-attention-layer-sequence}
]
def attn(x  : [S{s}, D], wq : [D, H1],
         wk : [D, H1]  , wv : [D, H2]) {  
  k : [S{s}, H1] = matmul(x, wk)
  v : [S{s}, H2] = matmul(x, wv)
  q : [S{s}, H1] = matmul(x, wq)
  qt: [H1, S{s}] = transpose(q)
  k_: [S, H1]    = all_gather {s} k
  a : [S, S{s}]  = matmul(k_, qt)
  b : [S{s}]     = reduce(*@$_{1,\mathtt{add}}$@*)(a)
  c : [S, S{s}]  = broadcast(*@$_0$@*)(b)
  d : [S, S{s}]  = div(a, c)
  z_: [S, H2]    = matmul(d, v)
  z : [S{s}, H2] = reduce_scatter {s} z_
  return z
}
\end{lstlisting}
%\vspace{-12pt}
\subcaption{One possible resolution for sequence sharding}
\label{fig:simplified-attention-layer-sequence}
\end{minipage}
\end{minipage}
\begin{minipage}[t]{0.57\linewidth}
%\vspace{0pt}
\begin{minipage}[t]{\linewidth}
\begin{lstlisting}[
    language=python,
    basicstyle=\ttfamily\footnotesize,
    label={lst:simplified-attention-layer-less-identification}
]
def attn(x  : [S, F]  , wq : [D1, H1],
         wk : [D2, H2], wv : [D3, H3]) {
  k : [S1, H21] = matmul(x : [S1, F1], wk : [F1, H21])
  v : [S2, H31] = matmul(x : [S2, F2], wv : [F2, H31])
  q : [S3, H11] = matmul(x : [S3, F3], wq : [F3, H11])
  qt: [H111, S31] = transpose(q : [S31, H111])
  a : [S11, S311] = matmul(k : [S11, H211], qt : [H211, S311])
  
  b : [S3111] = reduce(*@$_{1,\mathtt{add}}$@*)(a : [S111, S3111])
  c : [Sc, S31111] = broadcast(*@$_0$@*)(b : [S31111])
  d : [Sc1, S3112] = div(a : [Sc1, S3112], c : [Sc1, S3112])
  
  z : [Sc11, H311] = matmul(d : [Sc11, S21], v : [S21, H311])
  return z
}
\end{lstlisting}
%\vspace{-12pt}
\subcaption{\NDA results, after identifying with $\Isubst$ only}
\label{fig:simplified-attention-layer-less-identification}
\end{minipage}
\begin{minipage}[t]{\linewidth}
%\vspace{12pt}
\centering
\includegraphics[width=\linewidth]{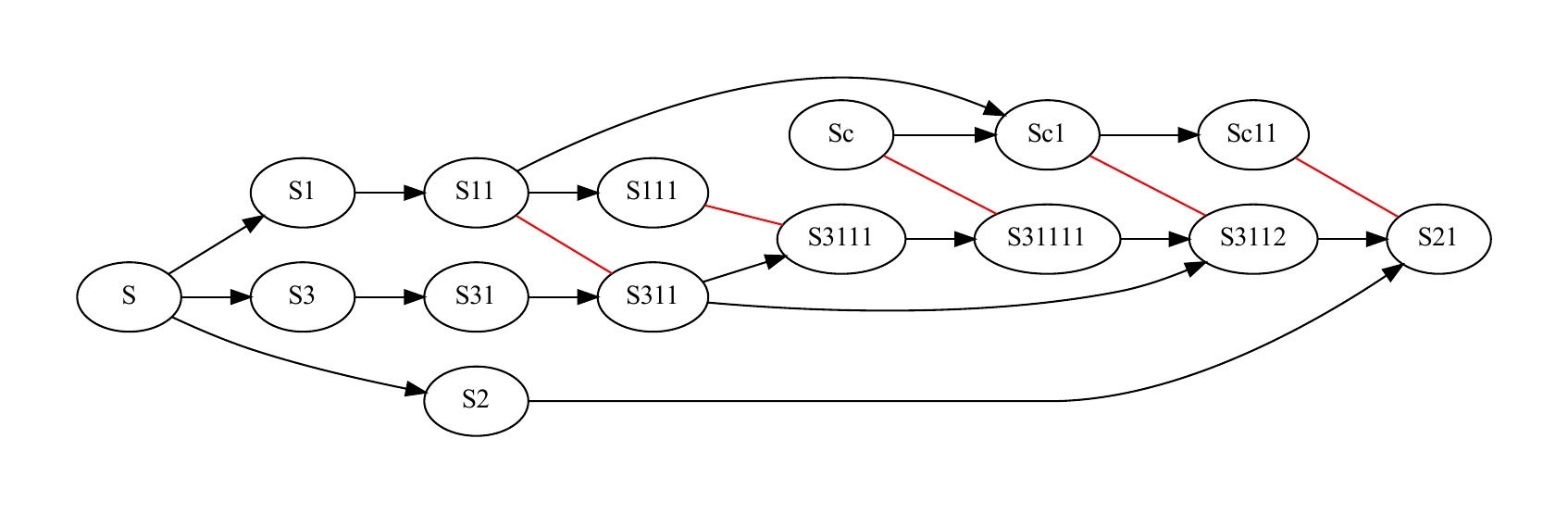}
%\vspace{-6pt}
\subcaption{Dimension graph (component connected of \texttt{S}): nodes refer to dimension \names in listing (c) above; conflicts shown as undirected edges (in red)}
\label{fig:simplified-attention-layer-dim-graph}
\end{minipage}
\end{minipage}
\caption{Simplified attention computation (mocking-up softmax as averaging for illustration purposes)}
\label{fig:simplified-attention-layer}
\end{figure*}

When the same dimension \name occurs more than once among the \names assigned to a tensor, we refer to this as a {\em sharding conflict} (or just {\em conflict}).
Conflicts prominently occur in the attention layer of the Transformer model \cite{DBLP:journals/corr/VaswaniSPUJGKP17}.
\Cref{fig:simplified-attention-layer-NDA} shows a simplified self-attention computation from the Transformer architecture, annotated with dimension \names after identifying with $\Isubst$ and $\Map$.
Up to the definition of \inlc{a}, the attention computation is standard.
Notice that \inlc{a} has a conflict since \texttt{S} occurs twice in its annotation.

To ease the presentation, \Cref{fig:simplified-attention-layer-NDA} replaces the \inlc{softmax} computation in standard attention with an averaging computation, which (like \inlc{softmax}) includes a reduction and pointwise operation.
% The full \inlc{softmax} computation includes an exponential function and, typically, another \inlc{reduce} and further \inlc{broadcast}s.
Note that the \inlc{reduce} makes the conflict from \inlc{a} disappear, since it removes the reduced-over dimension.
Conflicts appear again in the results of \inlc{broadcast} and \texttt{div}:
the \textsc{op} rule in \Cref{fig:NDA} forces the conflict from \inlc{a} also onto \inlc{c} and \inlc{d}.
Lastly, the final \inlc{matmul} contracts over dimension \texttt{S} and thus removes the conflict present in \inlc{d}.

When sharding a set of dimensions that are labeled with the same dimension \name, conflicts may generally need to be {\em resolved}.
That is, for each tensor where the dimension \name appears more than once, one must choose one of the dimensions for sharding.
In \Cref{fig:simplified-attention-layer-sequence}, the conflicts in the attention layer have been resolved by sharding the last dimensions of \inlc{a}, \inlc{c} and \inlc{d} along axis \texttt{s}, introducing an \texttt{all\_gather} and a \texttt{reduce\_scatter} operation. This partitioning of the \texttt{attn} function is known as {\em sequence sharding} \cite{seq_paralleism_nvidia} and is required for scaling the Transformer sequence length. 

% For each of these tensors one could resolve the conflict by picking one of the two dimensions labeled \inlc{S} for sharding.
% This results in a total of 8 different partitionings of \inlc{attn}, each corresponding to a different way of resolving the conflicts present in the function body.
% Only one of these 8 partitionings is the desirable sequence sharding from (TODO: cite Ref!!).

\subsection{Sharding in the presence of conflicts}
\label{sec:NDA:handling-conflicts}
Because of the possibility of conflicts, we cannot simply shard all dimensions that have been assigned the same \name by the \NDA (after identifying \names with $\Isubst$ and $\Map$).
We must additionally specify how conflicts should be resolved if they occur.
Instead of defining another analysis that discovers conflicts and offers resolutions, we can in fact re-use the \NDA to produce also conflict resolutions.
The key idea here is:
\begin{quote}
Do not identify dimension \names with the definition-to-use map $\Map$, but only using the sharding rules identities $\Isubst$.%
\end{quote}
%\footnote{This is why we use $\circeq$ for entries of $\Isubst$, to deliberately make them look like identities, but not so for the entries of $\Map$.}

\begin{comment}
We use results computed by the \NDA to facilitate automatic partitioning.
Hence, it is only natural to expect that we can use \NDA results to enable our automatic partitioner to resolve conflicts.
In particular, this is required for automatically discovering sequence sharding of Transformer models, as discussed in the previous section.

Once we have fully identified the dimension \names computed by the \NDA, with all identities from $\Isubst$ and $\Map$, as in \Cref{fig:simplified-attention-layer-NDA}, it is too late to get a good handle on individual conflicts.
The conflicting dimension \name \texttt{S} has spread through the program.
Hence, additional work, i.e.~analysis, would be required again to discover which tensors actually have conflicts (for \texttt{S}).

However, the \NDA has already computed this information.
We simply lost it when we identified dimension \names also with the entries of $\Map$.
\end{comment}

To illustrate this idea, we show the simplified attention layer from \Cref{fig:simplified-attention-layer-NDA} again in \Cref{fig:simplified-attention-layer-less-identification}, this time identifying dimension \names only with the identities in $\Isubst$.
As we have seen before, this means that each operation is annotated with fresh dimension \names, and each \name corresponds to precisely one way in which an operation can be partitioned.
Instead of listing the entries of the map $\Map$, a fraction of $\Map$ is pictured as a graph in \Cref{fig:simplified-attention-layer-dim-graph}:
each black, directed edge corresponds to an entry in $\Map$.
When we think of the full map $\Map$ as a graph, we refer to this as the {\em dimension graph}, since the nodes in this graph are dimension \names.
The fraction of the dimension graph in \Cref{fig:simplified-attention-layer-dim-graph} is the connected component of \texttt{S}.
This is the only connected component of $\Map$ whose dimension \names participate in conflicts.

When not identifying dimension \names with $\Map$, we must characterize conflicts differently from before:
a conflict occurs between {\em any} pair of dimension \names that annotate the same variable definition or use.
%
%We must include variable definitions {\em and} uses in this characterization because it is $\Map$ that ties together dimension \names from definitions and uses, but we are now not identifying dimension \names with $\Map$.
%
Conflicts in the \texttt{attn} function are drawn as red, undirected edges in \Cref{fig:simplified-attention-layer-dim-graph}.
Three of the five conflicts are precisely the conflicts in variables \inlc{a}, \inlc{c} and \inlc{d} that we saw in \Cref{fig:simplified-attention-layer-NDA} already.
The additional two conflicts come from the uses of \inlc{c} and \inlc{d}, respectively.

Each conflict edge can be resolved in two ways, picking one or the other endpoint -- a total of 32  resolutions.
% \dvytin{I deleted a ton of stuff below about dataflow propagation or propagation on color graphs because I revisit dataflow propagation alter in the agent discussion, and color graph propagation
% we have not even attempted or done. I think all the reader wants to know at this point is that they can shard all colors, carefully by resolving conflicts in given ways.}
%
Unfortunately, 32 resolutions for just sharding the sequence length of a transformer significantly % bumps up the space an agent would have to explore to pick a sharding,
increases the search space for automation.
Next, we discuss two heuristics for reducing the number of conflicts that need independent resolutions.

\begin{figure}
\begin{minipage}[t]{0.32\linewidth}
    %\vspace{0pt}
    \centering
    \includegraphics[width=\textwidth]{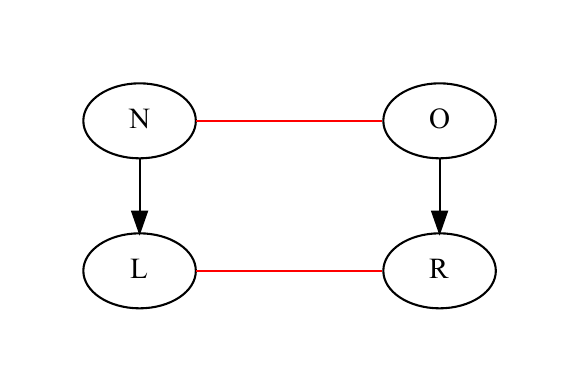}
\end{minipage}
\begin{minipage}[t]{0.32\linewidth}
    %\vspace{0pt}
    \centering
    \includegraphics[width=\textwidth]{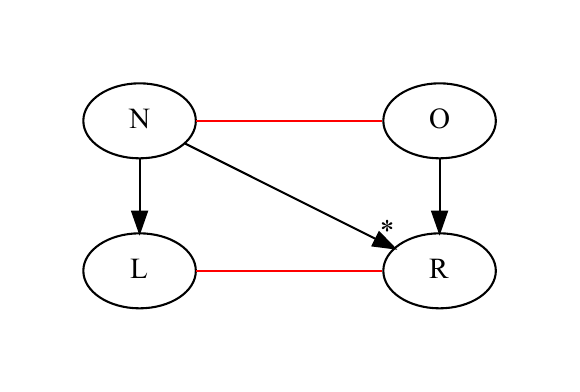}
\end{minipage}
\begin{minipage}[t]{0.32\linewidth}
    %\vspace{0pt}
    \centering
    \includegraphics[width=\textwidth]{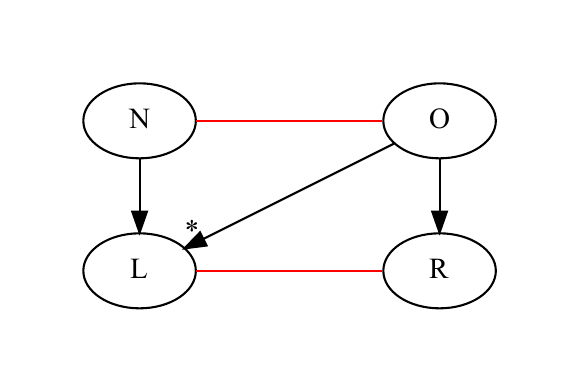}
\end{minipage}
\caption{``Box'' of compatible conflicts (left) and incompatible conflicts, with paths going ``across'' (middle and right).}
\label{fig:compatible-conflicts}
\end{figure}

\subsection{Compatible conflicts}
\label{sec:NDA:compatible-conflicts}
Conflicts in variable definitions and the corresponding variable uses should be attempted to be resolved in the same way, if we wish to avoid communication (resharding).
Consider a fragment of a dimension graph that looks like the ``box'' on the left of \Cref{fig:compatible-conflicts}:
\texttt{N} and \texttt{O} are dimension \names in the definition of a variable, and \texttt{L} and \texttt{R} are \names in a use of the same variable.
If the two conflicts are resolved differently, say \texttt{N} was sharded (but not \texttt{O}) and \texttt{R} was sharded (but not \texttt{L}), resharding with an \texttt{all\_to\_all} communication primitive \cite{MPI:1995} is required between variable definition and use.
On the other hand, if \texttt{N} and \texttt{L} are both sharded, no resharding (and hence no communication) is needed.

% We now explain two heuristics to reduce the number of possible conflict resolutions to a manageable number.
% To achieve this, we employ two ideas.
% \begin{enumerate}[(a)]
%     \item \label{item:conflicts:def-and-use}
%     Conflicts in variable definitions and the corresponding variable uses are effectively the same, and hence should be attempted to be resolved in the same way.
%     \item \label{item:conflicts:repeated-layers}
%     Conflicts in repeated layers of ML models should be resolved in the same way.
% \end{enumerate}

We refer to pairs of conflicts as {\em compatible} if, in the dimension graph, they form a ``box'' (\Cref{fig:compatible-conflicts}, left).
In real ML models, more complex dataflow may complicate matters:
there may also be paths in the dimension graph going across the ``box'' (\Cref{fig:compatible-conflicts}, middle and right).
If such paths exists, we do not deem conflicts compatible.

Motivated by avoiding extra communication while reducing the number of independent resolutions, we decree that compatible conflicts should be resolved in the same way.
We construct sets of compatible conflicts by considering the reflexive, symmetric and transitive closure of the compatibility relation, which we refer to as {\em compatibility sets}.
In \Cref{fig:simplified-attention-layer-dim-graph} there is only one compatibility set, containing all conflicts: \{%
(\texttt{S111}, \texttt{S311}),
(\texttt{S111}, \texttt{S31111}),
(\texttt{Sc}, \texttt{S31111}),
(\texttt{Sc1}, \texttt{S3112}),
(\texttt{Sc11}, \texttt{S21})\}.
There are two resolutions, one of which leads to the sequence sharding from \Cref{fig:simplified-attention-layer-sequence}.
The other introduces two {\tt all\_gather}s and has different memory and runtime characteristics; we omit it for lack of space.
% \dvytin{Should we say anything about the transitive closure being dangerous and having to be done carefully, or no space?}
% \nrink{We stress that what we present here is a heuristic. It does not need to be safe, so long as we can, in principle, detect unsafe situations and then handle them, e.g., by not applying the heuristic.}

\subsection{Conflict compatibility across layers}\label{sec:NDA:compatible-conflicts-layers}
Our second heuristic identifies compatibility sets of conflicts that originate from repeated layers in ML models.
Each compatibility set $C$ can be thought of as a sub-graph of the full dimension graph, containing just the nodes $C$ and any edges with endpoints in $C$.
For repeated layers, these graphs are necessarily isomorphic.
When compatibility sets $C_1$, $C_2$ %from different layers
are isomorphic, we decree that corresponding pairs of conflicts from $C_1$ and $C_2$ should be resolved in the same way.
% namely the subgraph of the dimension graph containing as nodes only those conflicting dimension \names from the set $C$.
% For repeated layers of ML models, these graph are necessarily isomorphic.
% We use the reverse statement as a heuristic that addresses point  \ref{item:conflicts:repeated-layers} above:
% When compatibility sets $C_1$, $C_2$ are isomorphic (as subgraphs of the dimension graph), we decree that corresponding pairs of conflicts from $C_1$ and $C_2$ should be resolved in the same way.

% TODO:
% Instead of checking graph isomorphic, we use the following simple proxy to determine when compatibility sets should be resolved identically.
For Transformers, this heuristic successfully identifies all (forward) attention layer compatibility sets, and also all corresponding compatibility sets in the backwards layers.%
\footnote{We omit the detailed example for reasons of space.}
Since each forward and backward layer has one compatibility set, 
we only get 4 different conflict resolutions for the Transformer architecture, regardless of the number of layers. 

\section{TOAST: The other auto-sharding tool}
\label{sec:toast}
\begin{figure}
    \centering
    % https://docs.google.com/drawings/d/13YSxJAOIGtl6MHl7_a-e_uFYhIYAcaooxJkxzuTD4Ag/edit?resourcekey=0-o5bQXUTYdAg6GxKMdY1f3g
    \includegraphics[width=\linewidth]{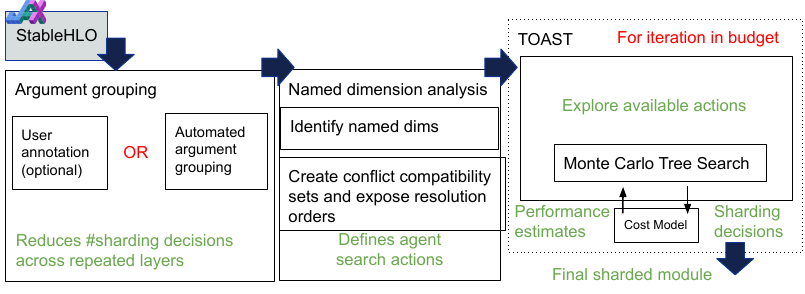}
    \caption{\label{fig:arch}\toast{} overall system architecture.}
\end{figure}

The \NDA exposes sets of tensor dimensions that should be sharded identically.
Importantly, it identifies conflict compatibility sets.
We now describe \toast{},  a system built to leverage the \NDA into an
automatic partitioner, via a simulated cost model for runtime and memory estimation interpreting device-local IR we generate.
% built as a simple abstract interpreter of device-local IR we generate. 
% (of the low-level device local IR we generate).
For reference, \Cref{fig:arch} tracks the various components of \toast{}. It is frontend (i.e., JAX~\cite{jax2018github}, PyTorch~\cite{pytorch2019}) and backend (e.g., XLA\cite{xla}) agnostic.
It operates over an array IR -- StableHLO~\cite{stablehlo, mlir2020},
that frontends frequently lower to, and backend compilers and runtime systems can ingest and execute on a variety of hardware platforms. The \toast{} partitioner is built as a Monte-Carlo Tree Search (MCTS)~\cite{Browne12asurvey}.
% conflicting colors, it will also surface all possible orders 

% The analysis described earlier expose set of tensors and their dimensions that should be sharded identically while avoiding needless communication.
% Which itself is very powerful - reducing the space of decisions a practitioner or a tool have to make to partition their program effectively. 
% The question of ~which~ dimensions to shard to provide an efficient partitioned module remains open. 
% Users want their partitioned modules to be fast subject to meeting on-device memory limits. 
% In this section we describe how we co-designed a search algorithm that leverages NDA to discover composition of different sharding strategies that achieve state-of-the-art (SOTA) performance.

% At a high-level, \toast{} is a search technique that uses the NDA to make decisions and identify the set of decisions it made so far. Then after making one or more number of sharding decisions, it queries a cost model that predicts the runtime and memory usage of the model after applying the sharding decisions on it, and feed that back to the search. 
% This runs in an iterative loop that can be stopped at any point by the user and get the best solutions discovered along the way.

% \input{sections/arch}

\subsection{MCTS: background}
\label{sec:toast:MCTS}
The goal of \toast{} is to find and apply sequences of actions that progressively shard an ML model. % and achieve the largest estimated improvements.
The core components are:
(1) the state $s$ of the search;
(2) the reward model $R(s)$ %that evaluates the state with an analytical cost model
(\Cref{sec:cost_model}) that assigns values to states, showing relative improvement (or deterioration) in expected performance;
(3) a set of actions that can be taken from a state $s$.

MCTS has been used successfully in many domains, including games, systems, and partitioning \cite{amer2013monte, silver2017mastering, neto2020multi, automap, automap-reloaded}. 
The algorithm balances exploration (gaining new information about possible action sequences) with exploitation (leveraging previously identified high-performing actions). 
Our implementation is multi-threaded, generating many trajectories, sequences of actions paired with their estimated performance, in parallel. 
A single round of MCTS consists of unrolling all trajectories and identifying the best-performing one to apply to the target module. 
Our method adapts a standard MCTS framework with several key heuristics tailored to our problem. 
First, while MCTS is typically run for a fixed computational budget, we terminate the entire search process early if a round of new trajectories fails to improve upon the best-known cost. 
Second, within the search, we weigh the MCTS actions to incentivize shorter trajectories. 
Early termination of individual rollouts is crucial because our system, \NDA, shards multiple dimensions simultaneously; shorter trajectories enable a more precise credit assignment, helping to better isolate and estimate the performance impact of sharding a specific dimension.

\subsection{Axis-aware and color-based actions}\label{mcts:actions}
Based on a model's \NDA and its compatibility sets ( \Cref{fig:simplified-attention-layer-dim-graph}), we design the available actions as tuples of the form:
\[      {\tt \dimname}~\times~{\tt resolution\_order}~\times~{\tt axis} \]
When an action is applied, the agent attempts to shard all dimensions corresponding to the ${\tt \dimname}$ onto the set of devices belonging to ${\tt axis}$. The ${\tt resolution\_order}$ is a bitstring to resolve the sharding conflicts that may arise.
Since each compatibility set offers two possible resolutions, a model with $b$ such sets requires a $b$-bit string for the ${\tt resolution\_order}$, where the $i$-th bit selects the resolution for the $i$-th set. 
In \toast{}, we pre-compute these compatibility sets at the construction time.
During the MCTS search, the agent's chosen action tuple determines which of the pre-calculated resolutions to apply. 
Each ${\tt \dimname}$ is encoded as a unique identifier, which we refer to as a \textit{color}.

% For each compatibility set we have two resolutions, and hence for the example in \Cref{fig:simplified-attention-layer-dim-graph}, with only one 
% compatibility set, we just need the resolution order to range in 0 or 1.
% In general, if there are $b$ compatibility sets $C_1,\ldots, C_b$, we need $b$ bits to express the {\tt resolution\_order}.
% When an action is applied, the agent attempts to shard all dimensions derived from this name. 
% Leveraging the co-design nature of \toast{}, we pre-compute these conflict sets at construction time and resolve them during the search using a predefined $\mathtt{resolution\_order}[i]$. To efficiently manage this process, each dimension name is encoded as a unique identifier, analogous to a color.

The MCTS search begins with a one-time setup where the initial action space is constructed from all possible triplets for the module. We prune this space by discarding actions affecting fewer than $10$ unique dimensions, as these trivial operations do not meaningfully improve performance.
The search then proceeds by simulating numerous trajectories from the game tree's root (unsharded module). 
Each trajectory simulation is a sequence of the following steps:
\begin{enumerate}
    \item An action is selected from the current state's set of available actions based on the MCTS selection policy.
    \item The simulation state is updated. This involves removing the action just taken from the available set, as well as pruning any other actions that have become invalid as a consequence of the new sharding state.
    \item The simulation of the trajectory ends, or reaches a terminal state, if one of the following conditions is met: (a) a special {\em stop} action is selected, or (b) a maximum trajectory depth is reached (which we set to $30$ actions).
\end{enumerate}
This process of simulating trajectories is repeated until the global computational budget is exhausted, at which point the best-found action sequence is returned.

\subsection{Colors aware state}
A key component of MCTS is the state representation, which must uniquely and efficiently identify each node in the search tree and its expected performance. 
A naive approach of tracking the sequence of applied actions is insufficient, as different action orderings can result in the same sharded model, leading to duplicated states and exponentially expands the amount of evaluation needed to explore the space. 
Another alternative, serializing the entire module after each action, while it is guaranteed to not introduce duplicates, is computationally prohibitive. 
A more recent approach \cite{automap-reloaded} tracks the sharding state of the function's arguments. 
However, this method cannot distinguish between different sharding configurations of intermediate tensors (i.e., tensors that are not function arguments). 
This ambiguity introduces a drawback where multiple distinct actions lead to the same state representation, necessitating complex, retrospective pruning mechanisms like {\em level 2 action transposition} \cite{actiongrouping}.

To overcome these limitations, we leverage \NDA to define our state. Our representation is an efficient in-memory map that records the sharding configuration of every dimension in the module. This design provides several key advantages:
\begin{itemize}
    \item \textbf{Efficient}: It avoids the cost of serializing the module by only tracking the dimensions that have been sharded.
    \item \textbf{Unambiguous}: The state is defined by the final sharding configuration itself, not the sequence of actions taken. Consequently, any action sequence yielding the same sharded model resolves to the same unique state, eliminating duplication by construction.
    \item \textbf{Simple}: By computing the action space ahead of time and exposing only unique, canonical actions, we eliminate the possibility of redundant actions in the search space. This proactive design avoids the need for complex run-time pruning mechanisms entirely.
\end{itemize}

\subsection{Grouping repeated layers}\label{sec:grouping}
% ML models consist of multiple layers, and each repeated layer effectively introduces its own set of (model) parameters.
% From a auto-sharding perspective, this exponentially increases the number of decisions.
% Conflict resolutions bring an additional exponential factor, which our heuristics from \Cref{sec:NDA:compatible-conflicts,sec:NDA:compatible-conflicts-layers} eliminates. This, however, only reduces the number of independent resolutions.
% It does not identify sharding decisions that should be made consistently across repeated layers.
The repeated layers common in modern ML models, each containing its own set of parameters, cause the decision space for auto-sharding to grow exponentially. Sharding conflicts introduce another layer of exponential complexity. While our heuristics (\Cref{sec:NDA:compatible-conflicts,sec:NDA:compatible-conflicts-layers}) address the latter by reducing the number of independent resolutions, they do not solve the initial problem. Specifically, they do not identify opportunities to apply consistent sharding decisions across these repeated layers.

To address this challenge, Alpa~\cite{alpa2022} discovers groups of sharding decisions that should be made consistently across model layers by timing and memory live range analysis.
\toast{}, on the other hand, uses a structural heuristic:
we group function arguments based on keys constructed from all uses of the (dimension \names of) these arguments.
The intuition is that repeated layers use function arguments that correspond to model parameters similarly.
Whenever an action is applied to a dimension, it is mirrored to the corresponding dimensions of all grouped arguments.

% Regardless of the grouping mechanism, once the model is annotated with groups, whenever an action is applied to a dimension (e.g., shard the contracting dimension), it is also mirrored to all grouped dimensions.
% \nrink{We can save space in this section by not talking about \toast{}'s manual grouping mechanisms. After all, we are presenting an automatic tool.}
% \nrink{If we focus on automatic mechanisms for grouping layers, then it is not worth mentioning Automap here. After all, this is not a Related Work section.}
% \nrink{Suggestion: leave the first paragraph in this section as it currently is, then finish off with the following:}
% \nr{
% To address this challenge, Alpa~\cite{alpa2022} discovers groups of sharding decisions that should be made consistently across model layers by (dynamic?) timing and memory liveness profiling.
% \toast{}, on the other hand, uses a static heuristic:
% we group function arguments based on keys constructed from all uses of the (dimension names of) these arguments.
% The intuition is that repeated layers use function arguments that correspond to model parameters similarly.
% Whenever an action is applied to a dimension, it is also mirrored to the corresponding dimensions of all grouped arguments.
% }

\subsection{Cost model}\label{sec:cost_model}
MCTS requires a way to evaluate the impact of a sequence of actions on a partitioned ML model. 
We rely on a fast and approximate cost model that involves analytical and roofline calculations to estimate compute op runtime as well as communication runtime for collectives (e.g. {\tt all\_gather}, {\tt all\_to\_all}, 
{\tt all\_reduce} etc). 
The cost model takes into account the device's characteristics, such as its FLOPS and network characteristics. 
The cost model is derived from an abstract interpreter of the MLIR module, where runtime cost is accumulated along the critical path. 
We take into account only matrix-multiplication ops (e.g.~{\tt dot\_general}, {\tt convolution}) and we provide cost estimates for different choices of collective implementation. 
We finally perform a live range analysis to approximate peak memory usage. 

Instead of absolute cost, MCTS only needs to know the {\em relative improvement} between states.
For example, we would expect batch partitioning across $b$ devices to reduce runtime by a factor of $b$.
% \footnote{For inference only. Training involves additional {\tt all\_reduce}s.}
Therefore, we define the cost of state $s$ as $C(s) = RT(s) + MP(s)$, where $RT(s)$ and $MP(s)$ are {\em relative} runtime ($RT$) and memory penalty ($MP$), respectively:
\begin{equation*}
\small
\begin{aligned}
    RT(s) &= \frac{\text{current runtime}}{\text{initial runtime}} \\
    MP(s) &= C\cdot \frac{\text{current peak} - DM}{\text{initial peak memory}} \cdot 
    \left\{\begin{array}{ll}
    1, & \text{if current peak} > DM \\
    0, & \text{otherwise}
    \end{array}\right.
    % \mathbf{1}_{\{\text{current peak} > DM\}}
\end{aligned}
\end{equation*}
$DM$ is the available amount of per-device memory.
So $MP$ penalizes a state $s$ only if the partitioned module's memory requirements (current peak) exceed the amount of local, per-device memory.
The constant $C$ controls how much of a penalty is incurred for exceeding memory constraints.
\section{Evaluation}\label{sec:eval}
\begin{figure*}
\centering
 \begin{subfigure}{0.18\linewidth}
        \centering
        \includegraphics[width=\linewidth]{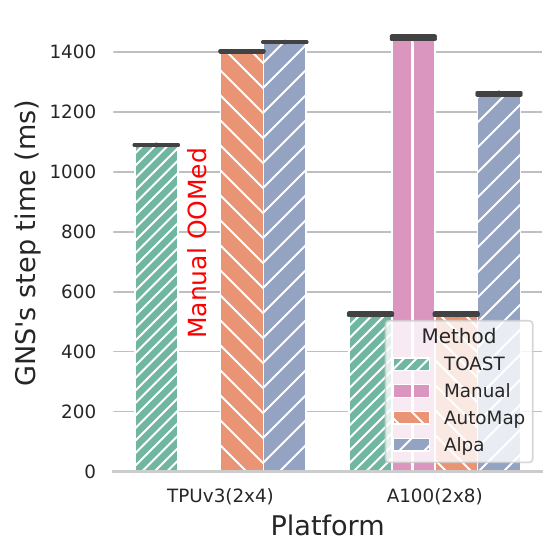}
        \caption{GNS}
        \label{fig:gns_step_time}
  \end{subfigure}
  \hspace{2pt}
  \begin{subfigure}{0.18\linewidth}
        \centering
        \includegraphics[width=\linewidth]{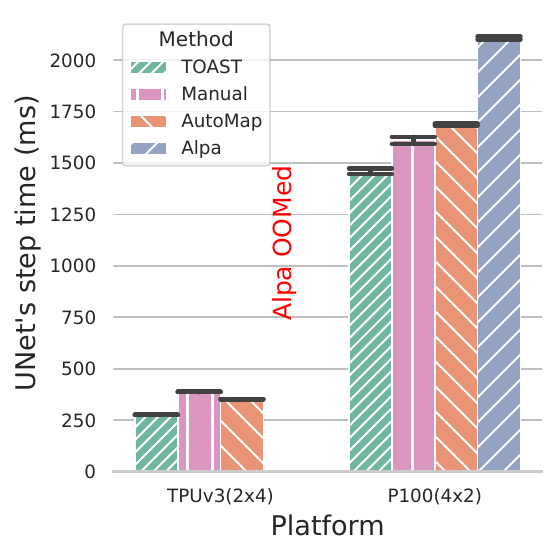}
        \caption{UNet}
        \label{fig:unet_step_time}
  \end{subfigure}
  \hspace{2pt}
  \begin{subfigure}{0.18\linewidth}
        \centering
        \includegraphics[width=\linewidth]{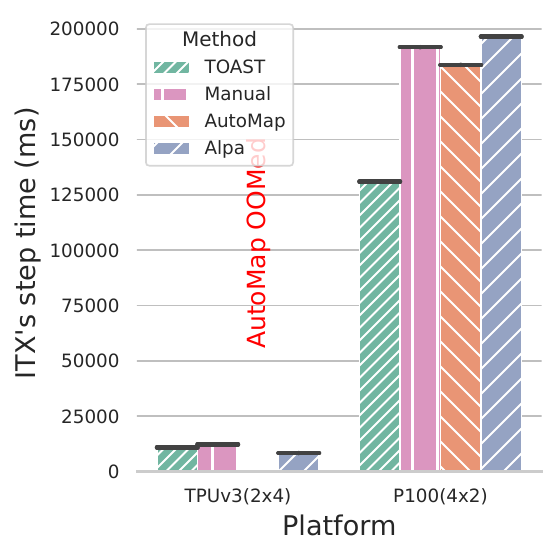}
        \caption{ITX}
        \label{fig:itx_step_time}
  \end{subfigure}
  \hspace{2pt}
  \begin{subfigure}{0.18\linewidth}
        \centering
        \includegraphics[width=\linewidth]{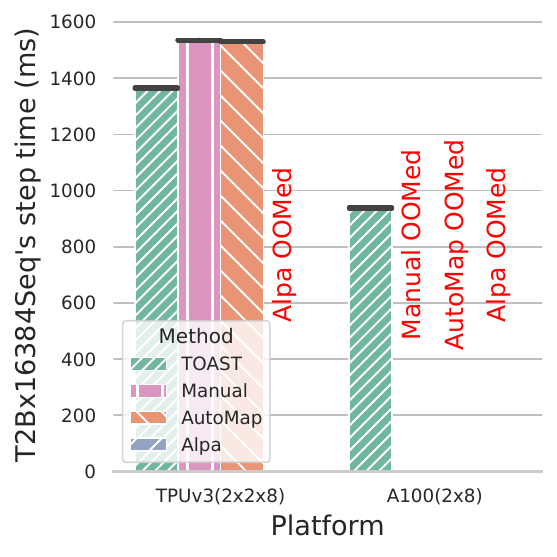}
        \caption{T2B 16kSeq}
        \label{fig:t2bx16384seq_step_time}
  \end{subfigure}
  \hspace{2pt}
  \begin{subfigure}{0.18\linewidth}
        \centering
        \includegraphics[width=\linewidth]{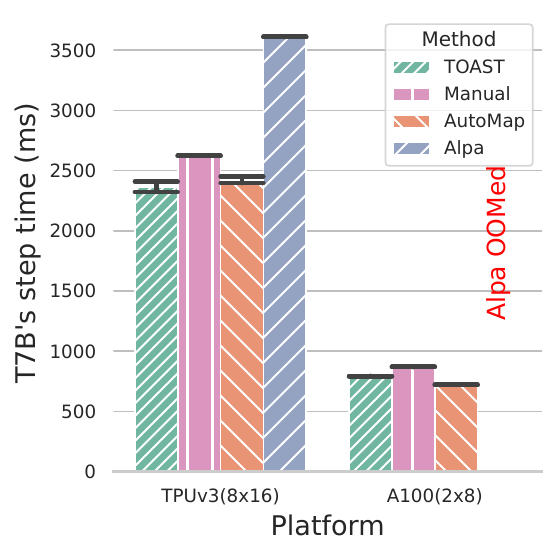}
        \caption{T7B 2kSeq}
        \label{fig:t7b_step_time}
  \end{subfigure}
\caption{Partitioned model step time in milliseconds (lower is better).}
\label{fig:step_time_all}
\end{figure*}

\begin{figure*}
\centering
  \begin{subfigure}{0.18\linewidth}
        \centering
        \includegraphics[width=\linewidth]{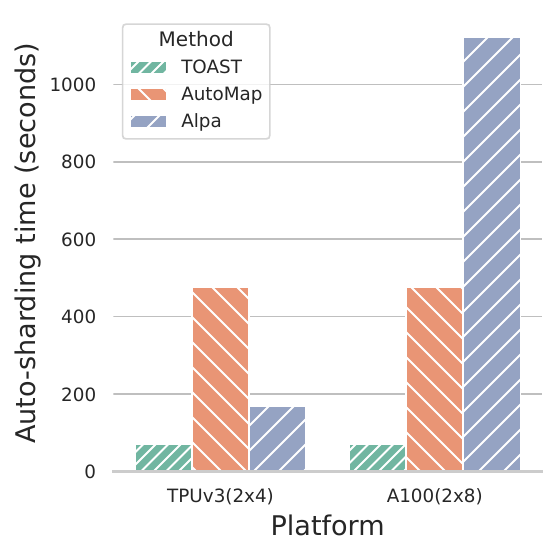}
        \caption{GNS}
        \label{fig:gns_search_time}
  \end{subfigure}
  \hspace{2pt}
  \begin{subfigure}{0.18\linewidth}
        \centering
        \includegraphics[width=\linewidth]{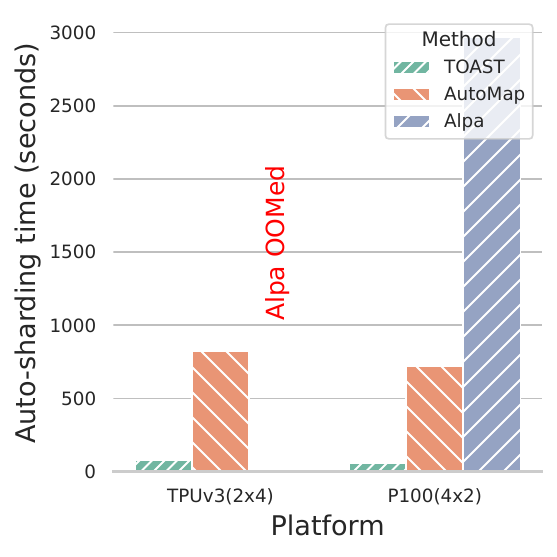}
        \caption{UNet}
        \label{fig:unet_search_time}
  \end{subfigure}
  \hspace{2pt}
  \begin{subfigure}{0.18\linewidth}
        \centering
        \includegraphics[width=\linewidth]{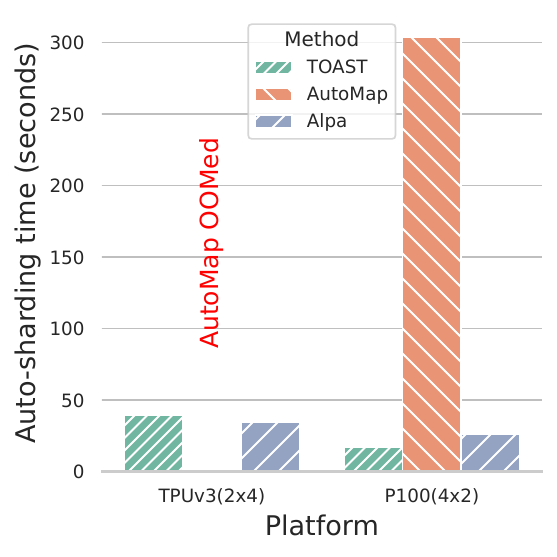}
        \caption{ITX}
        \label{fig:itx_search_time}
  \end{subfigure}
  \hspace{2pt}
  \begin{subfigure}{0.18\linewidth}
        \centering
        \includegraphics[width=\linewidth]{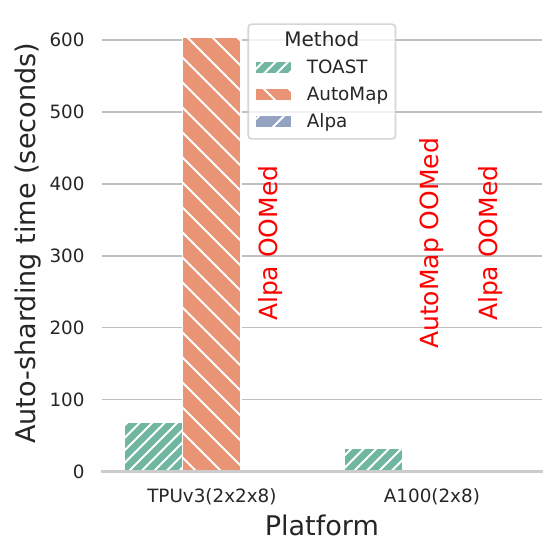}
        \caption{T2B 16kSeq}
        \label{fig:t2bx16384seq_search_time}
  \end{subfigure}
  \hspace{2pt}
  \begin{subfigure}{0.18\linewidth}
        \centering
        \includegraphics[width=\linewidth]{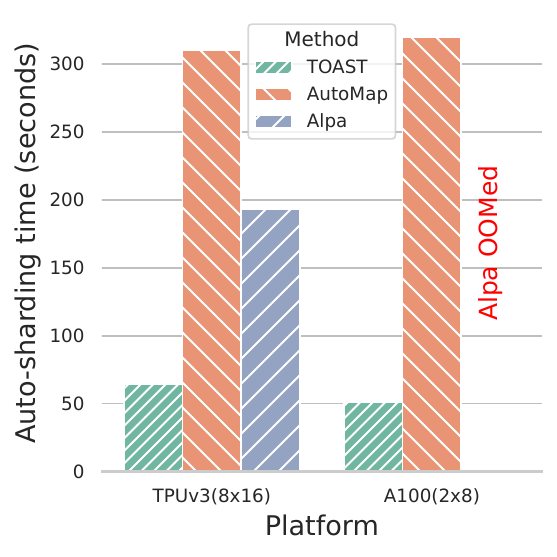}
        \caption{T7B 2kSeq}
        \label{fig:t7b_search_time}
  \end{subfigure}
\caption{Auto-sharding search time in seconds for the partitioned solution (lower is better).}
\label{fig:search_time_all}
\end{figure*}

We evaluate \toast{} on two criteria: (1) its ability to discover partitioning strategies that outperform expert-designed and related state-of-the-art (SOTA) methods across diverse hardware and model architectures (\Cref{eval:all}), and (2) its lower overhead compared to competing methods (\Cref{eval:overhead}).

\subsection{The setup}\label{eval:setup}
We partition a range of popular models used in competition \cite{benchmark_repo}. 
These models are written in JAX~\cite{jax2018github}, trained with Adam~\cite{adam}, and compiled with XLA~\cite{xla}.
We turned off XLA's rematerialization to avoid noisy measurements:
\begin{description}
\item [T2B/T7B] the 2B and 7B Gemma1 Transformer models~\cite{gemma1}:
\end{description}
{\centering
\footnotesize
\begin{tabular}{l|rrrrrr}
         & d$_{\text{model}}$ & layers & hidden dim. & heads & key size & vocab. \\
         \hline
     T2B & 2048 & 18 & 32768 &  8 & 256 & 256128 \\
     T7B & 3072 & 28 & 49152 & 16 & 256 & 256128
\end{tabular}
}

\noindent{\bf GNS} a 875M-parameters graph neural network used to simulate physical systems \cite{sanchez2020learning, godwin2022simple}.
Graphs contain 2048 nodes and 8192-65536 edges, mimicking a molecular structure.
We perform 24 message-passing steps over 3 linear layers (hidden size: 1024; latent size: 2048).

\noindent{\bf U-Net} a 3.6B-parameters convolutioncl neural network, popular in medical imaging \cite{og_unet, unet} and diffusion models~\cite{unet_diffusion}.
We use 9 residual down-sampling blocks, 12 up-sampling blocks, and between them a 32-head attention layer. % 

\noindent{\bf ITX} a 5B-parameters inference optimized transformer \cite{inference_transformer}, featuring ROPE \cite{rope} and a KV cache~\cite{inference_transformer}. 
(Vocab.~size: 50257;
sequence length: 1024;
prompt length: 1024;
32 heads;
32 layers;
hidden dim.: 4096;
d$\_{\text{model}}$: 2048.
) 

We conducted the experiments on set of A100~\cite{a100} and P100~\cite{p100} GPUs and on TPUv3~\cite{tpu_cloud}, with NVidia's NVLINK and ICI~\cite{tpuv4} connectivity respectively.
% Our acceralators are connected to different CPU servers that where the sharding and compiling is run, the A100s run on 2x Intel Cascadelake with 28 cores, the P100 run on 2x Intel Broadwell that have 22 cores, and TPUv3 
For the experiments, we partition the models with Alpa \cite{alpa2022}, AutoMap \cite{automap, automap-reloaded}, \toast{}, and expert-driven (i.e.~manual) annotations.

\subsubsection{Expert/manual sharding}
% For each of these models, we consider an expert-designed \textit{Manual} sharding strategy, which we detail below and use as a baseline in our figures:
For each model, we use an expert-designed \textit{Manual} sharding strategy as our baseline, which we detail below:
\begin{enumerate}
\item \textbf{T2B, T7B}: We partition these Transformer models using a fully-sharded data parallel (FSDP)~\cite{fsdp, zero2019} strategy, combined with Megatron-style~\cite{megatron2019} sharding of the MLP and attention-head layers, and sequence parallelism~\cite{seq_paralleism_nvidia}. 
Our \textit{Manual} baseline represents the best-performing configuration found after exhaustively searching all combinations of these strategies.
\item \textbf{GNS}: The SOTA strategy is edge sharding~\cite{edge_sharding}. However, we found that combining this with Megatron-style~\cite{megatron2019} partitioning of the linear layers within each node and edge processor of the GNS improves both runtime and memory performance.
\item \textbf{U-Net}: The industry-known \textit{manual} strategy combines FSDP~\cite{fsdp} with Megatron-style partitioning~\cite{megatron2019}.
\item \textbf{ITX}: For this inference-optimized Transformer model, the standard scaling approach combines multi-query attention sharding~\cite{inference_transformer}, Megatron partitioning~\cite{megatron2019}, and data parallelism over the batch dimension.
\end{enumerate}
Developing these manual strategies required significant expert effort to augment SOTA partitioning techniques with further sharding optimizations. 
The novelty and complexity of these approaches are evidenced by the multiple publications dedicated to them. 
We argue that \toast{}, despite being model- and hardware-agnostic, is highly competitive with these specialized, heavily-researched strategies.

\subsection{Auto-sharding of a wide range of models}\label{eval:all}
This set of experiments evaluates the generalizability of our automatic sharding tool across various models and hardware configurations.
A key motivation is that manual sharding annotations are often brittle; when a pre-trained model is adapted to a new use case or deployed on different hardware, its original annotations can lead to sub-optimal performance. 
Revising these annotations is a complex, error-prone process requiring significant expertise, which poses a substantial barrier to porting models effectively.

\Cref{fig:step_time_all} shows the {\em model step time}, the execution time for a single training or inference step—across all experiments, each repeated $10$ times for hundreds of steps. 
Across all tested models and hardware configurations, \toast{} consistently finds better partitioning strategies than the baselines. 
Notably, in cases where baselines do not cause an out-of-memory error (OOM), the step times of heavily optimized \textit{manual} strategies on TPUs, where they are battle-tested, are very close to those discovered by \toast{}. 
While Alpa and AutoMap have a cost model that penalizes solutions that exceed available on-device memory similar to \toast{}, in some scenarios (such as the 16k sequence length T2B experiment on GPU) they are not able to avoid OOMs, while \toast{} is able to do so. 
The action tuples available to \toast{} (\Cref{mcts:actions}) provide important trade-offs related to the order of conflict resolution, a key contribution of this paper which is unavailable to other automatic tools. 
By resolving conflicts in a specific order, \toast{} can reduce the peak memory usage of the partitioned model. 
This memory optimization is critical, as it can enable partitioning strategies with lower step times that would otherwise be infeasible due to memory constraints.
We want to stress that even a $1\%$ improvement in step time translates to weeks of saved computation, given that a model is compiled once with \toast{} but then trained or served for months at a time \cite{chinchilla2022}.

Another key result is that \toast{} consistently finds a superior sharding strategy across diverse hardware configurations, a crucial capability for enabling hardware fungibility at scale. 
Furthermore, \toast{}'s performance reveals that many model architectures are under-optimized. 
For instance, while its results on well-studied Transformer models are comparable to heavily tuned \textit{manual} strategies, it discovers significant improvements for other architectures. This finding suggests that standard industry techniques, such as the one used for GNS, are often suboptimal.

\subsection{Auto-sharding overhead}\label{eval:overhead}
A key differentiator among auto-sharding methods is the time required to find sharding decisions. 
An auto-sharding tool with a tolerable search time is more likely to be adopted for interactive model development, unlike methods suitable only for a model's final production run. 
\Cref{fig:search_time_all} compares the search times for Alpa, AutoMap, and \toast{}. 
We observe that Alpa, which is optimized for TPU workloads, is significantly slower on GPUs, while AutoMap and \toast{} exhibit platform-agnostic runtimes. 
We suspect this performance discrepancy stems from assumptions embedded within its linear programming solver; its cost model constraints are tuned for TPUs and thus require significantly more evaluation to meet GPU constraints.

While AutoMap is consistent with its performance across platforms, it remains significantly slower than \toast{}. 
AutoMap has to invoke the underlying propagation system after every action, which explains why, for complex models that need many actions, its search time can be orders of magnitude slower (25$\times$ for UNet and GNS). 
In contrast, \toast{} pre-computes the propagations ahead-of-partitioning. 
When the MCTS agent selects an action, it performs a fast, in-memory mutation to record the choice of color and resolution order. 
These decisions are only materialized and applied to the module during cost model estimation. 
Furthermore, many of the \NDA{} queries are heavily cached; once conflicts and their compatibility groups are computed, they are accessed in $O(1)$ time during the conflict resolution phase. 
These design choices make \toast{}'s search time largely platform- and model-size-agnostic.

\subsection{Scaling devices and sequence length}
\begin{figure}[htbp]
    \centering
    \begin{subfigure}{0.48\linewidth}
        \centering
        \includegraphics[width=\linewidth]{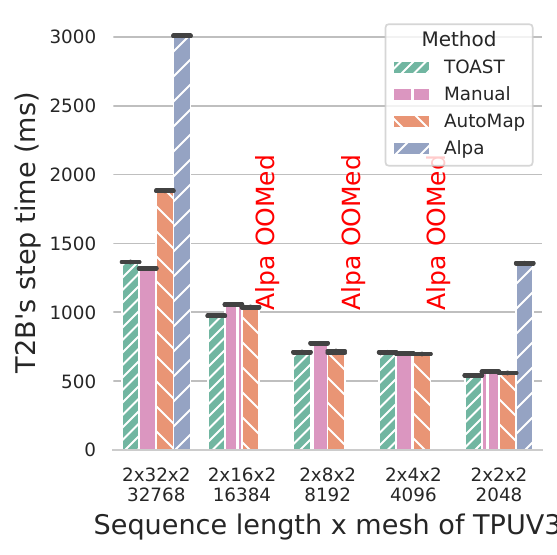}
        \caption{Step time in millisecond (lower is better).}
    \end{subfigure}
    \hspace{2pt}
    \begin{subfigure}{0.48\linewidth}
        \centering
        \includegraphics[width=\linewidth]{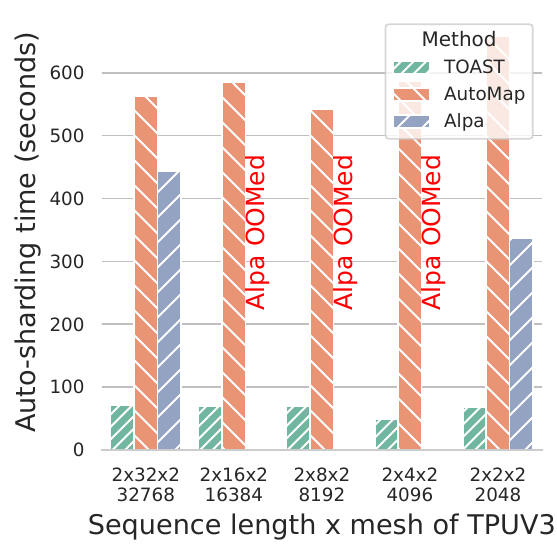}
        \caption{Search time in seconds (lower is better) scaling with number of devices.}
    \end{subfigure}
    \caption{\label{fig:scaling_with_big_compute}
    Partitioned T2B and scaling its sequence length and devices in a 3D mesh of `BatchxSeqxModel`, the number of devices can be calculated by multiplying the different axes size, e.g., $2x32x2 = 128$ device for the 32k sequence length.}
\end{figure}
Finally, we evaluate the scalability of our sharding solution as the model's sequence length and the number of devices increase. 
A good sharding solution must stay within the memory constraints of the available hardware and achieve performance comparable to an expert-derived baseline. 
Scaling a Transformer's sequence length is particularly important, as it enables it to capture the long-range dependencies between tokens to solve complex real-world problems.

As shown in \Cref{fig:scaling_with_big_compute}, \toast{} achieves near-optimal scaling, with performance comparable to expert-driven sharding strategies that required hundreds of engineering hours to develop. 
It outperforms SOTA methods like AutoMap by leveraging the novel conflict resolution and \NDA{} to operate over intermediate values, as motivated earlier in this paper. 
In contrast, under the high memory pressure of these scaling experiments, Alpa frequently triggered OOM errors or failed to find optimal solutions.
\section{Related work}
Partitioning systems like GSPMD~\cite{gspmd2021}, Shardy~\cite{shardy}, and PartIR~\cite{partir24_full} rely on user manual sharding. They are effective at propagating the manual sharding on inputs, but require users to know where to place the manual shardings. Without a clear signal how far the propagation will go, users are left with trial and error as the main mechanism to discover good sharding decisions. Propagation based on PartIR and GSPMD may introduce conflicts, that need to be handled with sharding constraints in the case of GSPMD, and special tag ops in PartIR case.
Discovering where to place these constraints requires often a lengthy iteration and profiling by experts. 
Our \NDA and conflict resolution system identifies a small set of these conflicts ahead-of-time, and serializes their resolution in the action space.
Fully-automatic partitioners such as Alpa\cite{alpa2022}, FlexFlow\cite{flexflow2019}, Unity\cite{unity2022}, and Galvatron\cite{galvatron} developed to simplify the task of sharding for users, using a range of
techniques. For example Alpa is based on constraint solving, with current work attempting to more iteratively expose the search space (\`a la~\cite{automap, automap-reloaded})

\section{Conclusion}
We have presented \toast{}, a fully automatic partitioning system that leverages our novel \named dimension analysis (\NDA) to expose partitioning decisions to an MCTS agent, 
including the exploration or resolutions of sharding conflicts. 
We show by providing ahead-of-time propagation decisions and operating over the \NDA our automatic partitioner is able to explore more performant solutions and achieve better results than other competing methods.

For future research, we will leverage \NDA{} to train a model capable of predicting the optimal sharding immediately given set of colors to shard. This is achievable by training offline on many combinations of colors as the cost estimate from sharding sets of \named dimensions should be deterministic.

\bibliographystyle{ACM-Reference-Format}
\bibliography{main}

% \newpage
% \appendix 
% \onecolumn % Revert back to 1 column style to make proofs readable
% \input{appendix/extended_eval}
% \input{appendix/proof}

\end{document}